\documentclass[lettersize,journal]{IEEEtran}
\usepackage{amsmath,amsfonts}
\usepackage{array}
\usepackage[caption=false,font=normalsize,labelfont=sf,textfont=sf]{subfig}
\usepackage{textcomp}
\usepackage{stfloats}
\usepackage{url}
\usepackage{verbatim}
\usepackage{graphicx}
\usepackage{cite}
\usepackage{xcolor}
\usepackage[ruled]{algorithm2e}
\usepackage{algpseudocode}
\usepackage{multirow}
\usepackage{bbding}
\usepackage{xcolor}
\definecolor{mygray}{gray}{.9}
\usepackage[capitalize]{cleveref}

\usepackage{amssymb}

\DeclareMathOperator*{\argmax}{arg\,max}
\DeclareMathOperator*{\argmin}{arg\,min}
\newcommand{\revisionb}[1]{\textcolor{black}{#1}}
\newcommand{\revision}[1]{\textcolor{black}{#1}}

\begin{document}

\title{Vulnerabilities in AI-generated Image Detection: \\The Challenge of Adversarial Attacks }

\author{Yunfeng Diao$^*$,~\IEEEmembership{Member,~IEEE}, Naixin Zhai$^*$, Changtao Miao, Zitong Yu,~\IEEEmembership{Senior Member,~IEEE}, Xingxing Wei, \\Xun Yang$^\dagger$,~\IEEEmembership{Senior Member,~IEEE}, Meng Wang,~\IEEEmembership{Fellow,~IEEE}

\thanks{$^*$ Yunfeng Diao and Naixin Zhai contributed equally to this paper.}
\thanks{$^\dagger$ Xun Yang is the corresponding author.}
\thanks{Yunfeng Diao and Meng Wang are with the Hefei University of Technology, Hefei, China (email:diaoyunfeng@hfut.edu.cn, eric.mengwang@gmail.com). Yunfeng Diao is also with Intelligent Interconnected Systems Laboratory of Anhui Province (Hefei University of Technology) .}
\thanks{Naixin Zhai, Changtao Miao and Xun Yang are with the University of Science and Technology of China, Hefei, China (email:\{zhainaixin,miaoct\}@mail.ustc.edu.cn, xyang21@ustc.edu.cn).} 
\thanks{Zitong Yu is with the School of Computing and Information Technology, Great Bay University,  Dongguan, China (e-mail: yuzitong@gbu.edu.cn).}
\thanks{XingXing Wei is with the Institute of Artificial Intelligence, Beihang University, Beijing, China (e-mail: xxwei@buaa.edu.cn).}}

\markboth{Journal of \LaTeX\ Class Files,~Vol.~14, No.~8, August~2021}%
{Shell \MakeLowercase{\textit{et al.}}: A Sample Article Using IEEEtran.cls for IEEE Journals}

\maketitle

\begin{abstract}
Recent advancements in image synthesis, particularly with the advent of GAN and Diffusion models, have amplified public concerns regarding the dissemination of disinformation. To address such concerns, numerous AI-generated Image (AIGI) Detectors have been proposed and achieved promising performance in identifying fake images. However, there still lacks a systematic understanding of the adversarial robustness of AIGI detectors. In this paper, we examine the vulnerability of state-of-the-art AIGI detectors against adversarial attack under white-box and black-box settings, which has been rarely investigated so far. To this end, we propose a new method to attack AIGI detectors. First, inspired by the obvious difference between real images and fake images in the frequency domain, we add perturbations under the frequency domain to push the image away from its original frequency distribution. Second, we explore the full posterior distribution of the surrogate model to further narrow this gap between heterogeneous AIGI detectors, e.g., transferring adversarial examples across CNNs and ViTs. This is achieved by introducing a novel post-train Bayesian strategy that turns a single surrogate into a Bayesian one, capable of simulating diverse victim models using one pre-trained surrogate, without the need for re-training. We name our method as Frequency-based Post-train Bayesian Attack, or FPBA. Through FPBA, we demonstrate that adversarial attacks pose a real threat to AIGI detectors. FPBA can deliver successful black-box attacks across various detectors, generators, defense methods, and even evade cross-generator and compressed image detection, which are crucial real-world detection scenarios. Our code is available at \url{https://github.com/onotoa/fpba}.
\end{abstract}


\section{Introduction}

\begin{figure}[!htb]
  \centering
  \includegraphics[width=1\linewidth]{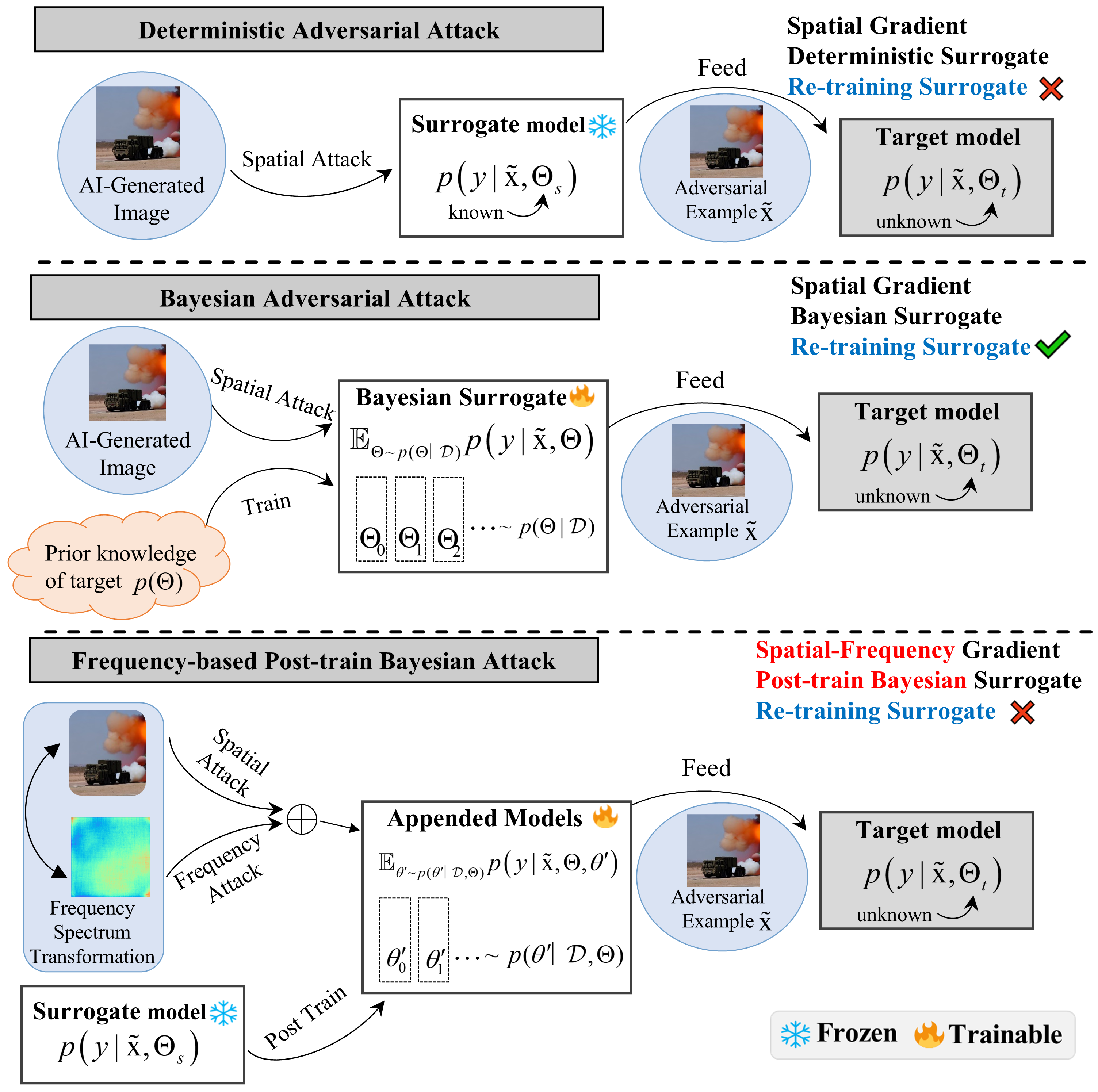}
  \vspace{-0.6cm}
  \caption{A high-level illustration of our proposed method.}
\label{fig:high_level}
\vspace{-0.6cm}
\end{figure}

The notable progress in generative models, such as GANs~\cite{ffhq} and Diffusion models~\cite{adm}, is driving the flourishing development of the image synthesis domain. These generated fake images exhibit realistic-looking, rendering them visually indistinguishable from real images. Moreover, a variety of free and open-source tools facilitates the effortless creation of fake images. However, alongside the benefits, the widespread availability of fake images raises concerns regarding the dissemination of misinformation and fake media news. 

Consequently, numerous detectors have been proposed to identify AI-generated images (AIGI). Recent state-of-the-art detectors~\cite{wang2020cnn, zhong2023rich} rely on Deep Neural Networks (DNNs) to classify, achieving significant accuracy performance across multiple datasets and generative models. However, our investigation reveals that AIGI detectors are vulnerable to adversarial examples, capable of misleading detectors by classifying fake images as real. \revision{Albeit identifying a key issue that needs to be addressed, designing an effective attack for AIGI detection is still challenging. Unlike image classification, which primarily investigates adversarial vulnerability in high-level semantic representations shared across models~\cite{I-FGSM,luo2022frequency}, AIGI detection relies on discriminative information from both high-level semantics and subtle low-level generative artifacts. This combination produces a more diverse and heterogeneous feature space, where adversarial perturbations must simultaneously disrupt semantic cues and model-specific generative fingerprints. Consequently, transferring adversarial patterns across such diverse representations is far less straightforward. Next, several works have explored the adversarial robustness in GAN-based Deepfake detection~\cite{carlini2020evading,hussain2021adversarial,neekhara2021adversarial,hou2023evading,jia2022exploring}. These methods largely focus on identifying facial fingerprints unique to GANs and exploiting real-fake differences in manipulated human faces. In contrast, AIGI detectors are designed to identify any AI-generated content produced by a wide range of generative techniques, including GANs, diffusion models, and autoregressive models. The content extends beyond human faces to encompass non-human entities, landscapes, abstract art, and more. These differences make designing attacks for AIGI detection more challenging than Deepfake detection.}

In this paper, we show that AIGI detectors are vulnerable to adversarial attacks. Considering that many works~\cite{dzanic2020fourier,frank2020leveraging} have demonstrated the obvious changes between real and fake images in the frequency domain, we explore the vulnerable region of AIGI detectors in the frequency domain. As illustrated in \cref{fig:vis_ssm}, the frequency components that different detectors focus on significantly vary from each other. Therefore, we utilize frequency spectrum transformation to uncover diverse substitute models via adding adversarial perturbations in the various frequency transformation domains. Further, Transformer-based detectors have demonstrated outstanding performance in detecting AI-generated images\cite{zhu2024genimage}, but we have observed that there is an obvious gap in adversarial transferability across heterogeneous AIGI detectors, e.g., transferring adversarial examples across Convolutional Neural Networks (CNNs) to Visual Transformers (ViTs) (as shown in \cref{tab:commands}). To tackle this issue, we propose a post-train Bayesian strategy to conduct a Bayesian treatment on the surrogate model, without the need for re-training the surrogate. In contrast to existing ensemble-based or Bayesian attacks, which involve retraining an ensemble of surrogate models, our post-train Bayesian strategy freezes the pre-trained surrogate and appends tiny extra Bayesian components behind the surrogate, avoiding a heavy memory footprint and speeding up the training process. As a result, we propose a new transferable adversarial attack for general AIGI detection, to add adversarial perturbations in various frequency transformation domains from a post-train Bayesian perspective. We name our method Frequency-based Post-train Bayesian Attack, or FPBA. A high-level illustration of our method and the key differences between our method and the previous method are shown in \cref{fig:high_level}.

The contributions of this work can be summarized as follows: (1) We systematically assess the adversarial robustness of state-of-the-art AIGI detectors, revealing both real-world threats and gradient masking effects. (2) We propose a new attack against AIGI detection by exploring the vulnerable frequency region using a post-train Bayesian strategy, avoiding retraining the surrogate model. (3) Extensive experiments across 17 AIGI detectors show that our method achieves the highest attack success rates under white-box and black-box settings, outperforming existing baselines.

\section{Related Work}

\noindent{\textbf{AI-Generated Image Detection: }}
The rapid progress of generative models, from early GANs \cite{ffhq} to recent diffusion models \cite{adm}, has raised substantial risks of disinformation. While early work focused primarily on detecting fake human faces \cite{miao2025multi,miao2025mixture}, the versatility of diffusion models has broadened generation to diverse scenes, posing greater challenges for detection. AIGI detection is typically formulated as a binary classification task to distinguish real from synthetic images. Data-driven methods \cite{wang2020cnn,feng2025deepfake} achieve strong in-distribution performance but generalize poorly to unseen generators. To address this, frequency-domain approaches \cite{miao2022hierarchical,miao2023f} exploit forgery traces, while spatial-domain methods learn local forgery features \cite{guo2023ldfnet}. Others leverage fingerprint representations from noise patterns \cite{liu2022detecting}. Parameter-efficient strategies based on frozen pre-trained models have also been explored \cite{tan2023learning}, along with approaches exploiting diffusion reconstruction error \cite{ricker2024aeroblade}, teacher–student discrepancies \cite{zhu2023gendet}, or diffusion noise \cite{zhang2023diffusion}.

\noindent{\textbf{Adversarial Attack: }}
Adversarial attacks craft perturbed examples that mislead target models, raising concerns in safety-critical applications such as image classification \cite{ran2024adaptive,I-FGSM}, multimedia communication \cite{gao2024deepspoof}, computational imaging~\cite{liang2025understanding} and human activity recognition \cite{diao2024understanding,diaotasar,diao2021basar}. \revision{Gradient-based attack methods~\cite{I-FGSM,pgd} exploit gradients to mislead models. To improve adversarial transferability, frequency-domain approaches~\cite{luo2022frequency} perturb spectral representations, while ensemble-based attacks~\cite{dong2018boosting,svre} aggregate gradients or perturbations from multiple models.} Very recently, their impact on Deepfake detection has been explored. Gradient sign-based attacks \cite{hussain2021adversarial,I-FGSM,fgsm}, latent space manipulations \cite{li2021exploring}, and black-box evaluations \cite{carlini2020evading,neekhara2021adversarial} reveal vulnerabilities in existing detectors. Other works leverage frequency-domain perturbations \cite{jia2022exploring}, natural degradation noise \cite{hou2023evading}, or fingerprint removal strategies such as FakePolisher \cite{huang2020fakepolisher} and TraceEvader \cite{wu2024traceevader}. Unlike Deepfakes, which primarily manipulate real face images, AI-generated images encompass far more diverse synthetic content, amplifying disinformation risks~\cite{zhou2024stealthdiffusion}. In this work, we propose a universal attack against both AIGI and Deepfake detectors, evaluating under both white-box and practical black-box settings.

\section{Methodology}
\begin{figure*}[!htb]
  \centering
  \includegraphics[width=0.8\linewidth]{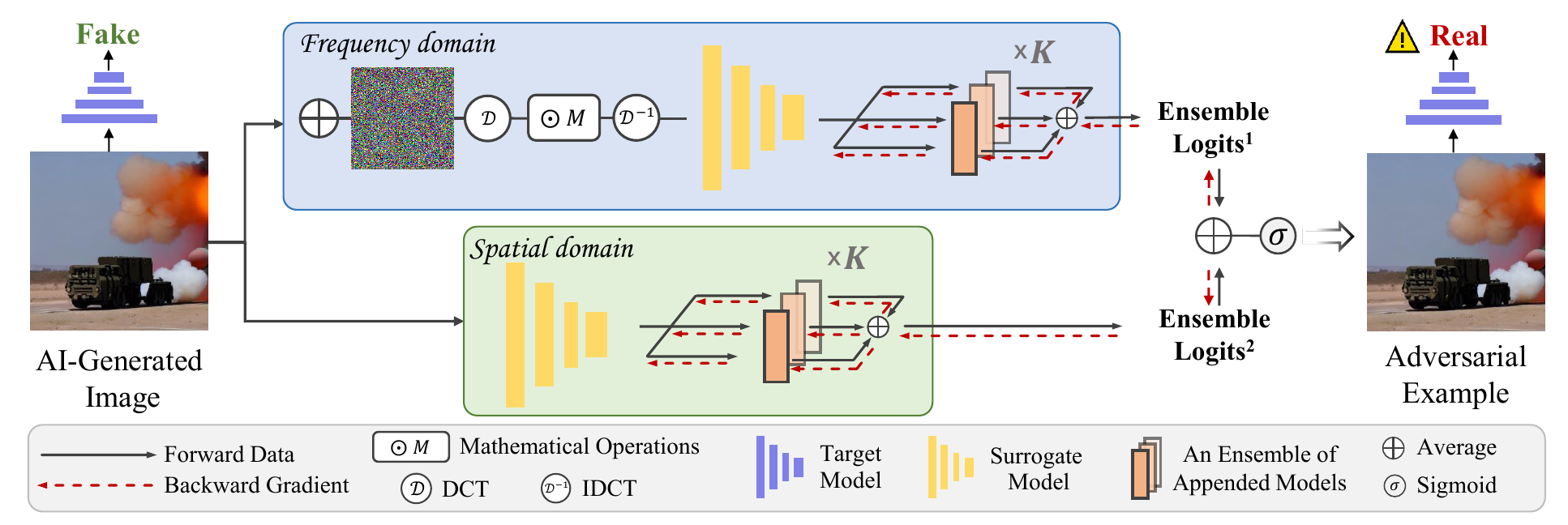}
  \vspace{-0.3cm}
  \caption{The workflow of FPBA. We add spatial-frequency adversarial perturbations to AI-generated images in a Bayesian manner, so that they are misclassified as real. DCT and IDCT are the discrete cosine transformation and inverse discrete cosine transformation, respectively.}
  \label{fig:overview}
  \vspace{-0.5cm}
\end{figure*}

\subsection{Preliminaries}
Let $\mathbf{x}$ and $y$ represent the original image and its corresponding label. $f_\Theta$ denotes the AI-generated image detectors. We aim to inject adversarial perturbation into the original image that makes the detector misclassify. Such an adversary problem can be optimized by minimizing the predictive probability, i.e., maximizing the classification loss:
\begin{equation}
\label{eq:1}
\argmin_{\Tilde{\mathbf{x}}} p(y \mid \Tilde{\mathbf{x}}, \Theta) 
= \argmax_{\Tilde{\mathbf{x}}} L(\Tilde{\mathbf{x}}, y, \Theta), \text { s.t. } \left\|\delta \right\|_{p} \leq \epsilon,
\end{equation}
where $L$ is the binary cross-entropy loss in AI-generated image detection. Adversarial example $\Tilde{\mathbf{x}} =\mathbf{x} + \delta$, in which $\delta$ is the adversarial perturbation and $\epsilon$ is the perturbation budget. \cref{eq:1} can be performed with iterative gradient-based methods, such as PGD~\cite{pgd} or I-FGSM~\cite{I-FGSM}:
\label{eq:pgd}
\begin{align}
 \Tilde{\mathbf{x}}^{i+1}&=\Tilde{\mathbf{x}}^{i} + \alpha \cdot \operatorname{sign}(\nabla L\left(\Tilde{\mathbf{x}}^{i}, y, \Theta \right)) \nonumber \\ 
 &= \Tilde{\mathbf{x}}^{i} - \alpha \cdot \operatorname{sign}(\nabla \, log \, p(y \mid \Tilde{\mathbf{x}}^{i}, \Theta)).    
\end{align}

\begin{figure}[!htb]
  \centering
  \includegraphics[width=1\linewidth]{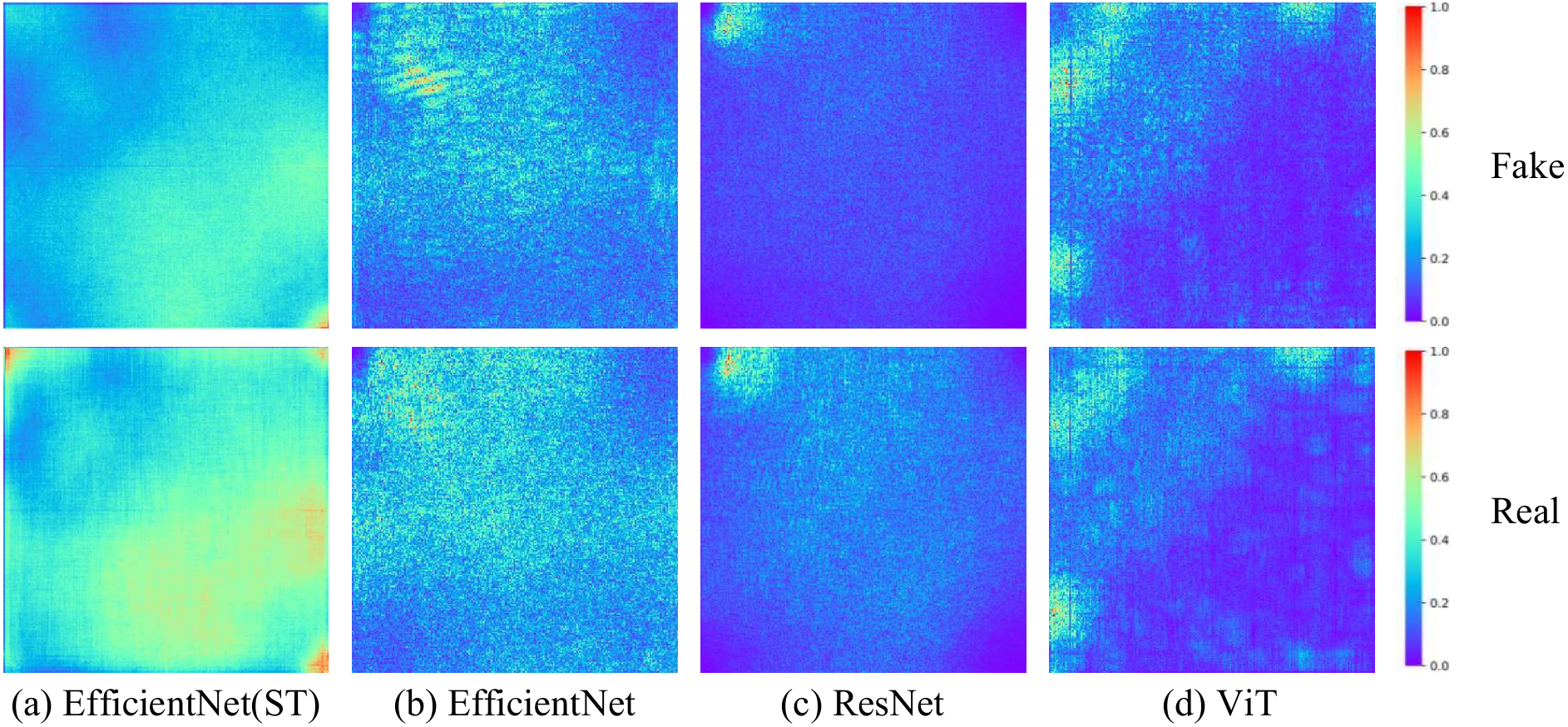}
\vspace{-0.6cm}
  \caption{
Visualization of the spectrum saliency map (average 2000 images on GenImage datasets) for real and fake images across different models. (a): the results for conducting frequency spectrum transformation (N=10). (b$\sim$d): the results for raw images on different models. The color value represents the absolute gradient value of the model loss function after max-min normalization.
}
\label{fig:vis_ssm}
\vspace{-0.5cm}
\end{figure}

\subsection{Frequency-based Analysis and Attacks}
\label{sec:intro}

Many AIGI detection approaches distinguish between real and fake images via subtle artifacts~\cite{wang2020cnn,zhong2023rich}. While these subtle clues are invisible in the spatial domain, a series of works~\cite{frank2020leveraging} demonstrate that there are obvious differences between real and fake images in the frequency domain. This inspires us to explore the vulnerable region of AIGI detectors from a frequency perspective. To this end, we first implement the discrete cosine transform (DCT) $\mathcal{D}(\cdot)$ to transfer the inputs from the spatial domain to the frequency domain. To investigate the difference between real images and fake images in the frequency domain, we use the spectrum saliency map~\cite{long2022frequency} to visualize the sensitive components of real and fake images across different models:  
\revision{
    \begin{equation}   
    \label{eq:ssm}
    \mathbf{S}_{\Theta} =  \frac{\partial J(\mathcal{D}^{-1}(\mathcal{D}(\mathbf{x}), y, \Theta)}{\partial\mathcal{D}(\mathbf{x})},
    \end{equation}}
where \revision{$\mathcal{D}^{-1}(\cdot)$} is the inverse discrete cosine transform (IDCT). In a spectrum map, the low-frequency components whose amplitudes are mainly distributed in the upper left corner, and the high-frequency components are located in the lower right corners. As shown in~\cref{fig:vis_ssm}, (1) There are significant differences between synthetic images and real images in the frequency domain. Therefore, moving the image away from its original frequency distribution will make the detectors hardly classify it as the ground-truth class. This observation motivates us to attack under the frequency domain to push the original images away from their ground-truth frequency distribution. (2) Different models usually focus on different frequency components for classifying (\cref{fig:vis_ssm}(b$\sim$d)). This inspires us to conduct random spectrum transformation to stimulate diverse substitute models. Followed by \cite{long2022frequency}, the spectrum transformation $\Gamma(\mathbf{x})$ is defined as:
\revision{
\begin{equation}
     \Gamma(\mathbf{x}) = \mathcal{D}^{-1}(\mathcal{D}(\mathbf{x} + \xi) \odot \mathcal{M}),
 \label{eq:st}
\end{equation}
}
where $\Gamma(\cdot)$ denotes the random spectrum transformation~\cite{long2022frequency}. $\odot$ is the Hadamard product, $\xi$ is a random noise drawn from an isotropic Gaussian $\mathcal{N}\left(0, \sigma^{2} \mathbf{I}\right)$, and each element of $\mathcal{M}$ is sampled from a Uniform distribution $\mathcal{U}(1-p, 1+p)$. As shown in \cref{fig:vis_ssm}(a), tuning the spectrum saliency map can cover most of the other models. We hence conduct adversarial attacks in the frequency domain via spectrum transformation:

\begin{align}
\label{eq:fre_attack}
    \argmin_{\Tilde{\mathbf{x}}} p(y \mid \Gamma(\Tilde{\mathbf{x}}), \Theta), \text { s.t. } \left\|\delta \right\|_{p} \leq \epsilon.
\end{align}

\subsection{Exploring the Surrogate Posterior Space}
Although tuning the spectrum transformation in \cref{eq:fre_attack} can simulate different substitute models with a homogeneous architecture~\cite{long2022frequency}, it shows limited transferability when applied to heterogeneous architectures, e.g., transferring adversarial examples across ViTs and CNNs. This motivates us to consider the frequency-based attack from a Bayesian perspective, i.e., exploring the full posterior distribution of the surrogate model to further narrow this gap between heterogeneous models. Therefore, we redefine \cref{eq:fre_attack} by 
minimizing the Bayesian posterior predictive distribution: 
\begin{align}
     &\argmin_{\Tilde{\mathbf{x}}} \, p(y \mid \Gamma(\Tilde{\mathbf{x}}), \mathcal{D}) \nonumber \\ 
     = &\argmin_{\Tilde{\mathbf{x}}} \, \mathbb{E}_{\Theta \sim p(\Theta\mid \mathcal{D})} p\left(y \mid \Gamma(\Tilde{\mathbf{x}}), \Theta\right) , \text { s.t. } \left\|\delta \right\|_{p} \leq \epsilon,
  \label{eq:Bay_attack}
\end{align}
where $p(\Theta \mid \mathcal{D}) \propto p(\mathcal{D} \mid \Theta) p(\Theta)$. $\mathcal{D}$ is the dataset and $p(\Theta)$ is the prior of model weights. Attacking Bayesian Neural Networks (BNNs) rather than a single DNN allows for the output fusion from an ensemble of infinitely many DNNs with diverse predictions, thereby improving adversarial transferability.
\begin{figure}[t]
  \centering
  \includegraphics[width=1\linewidth]{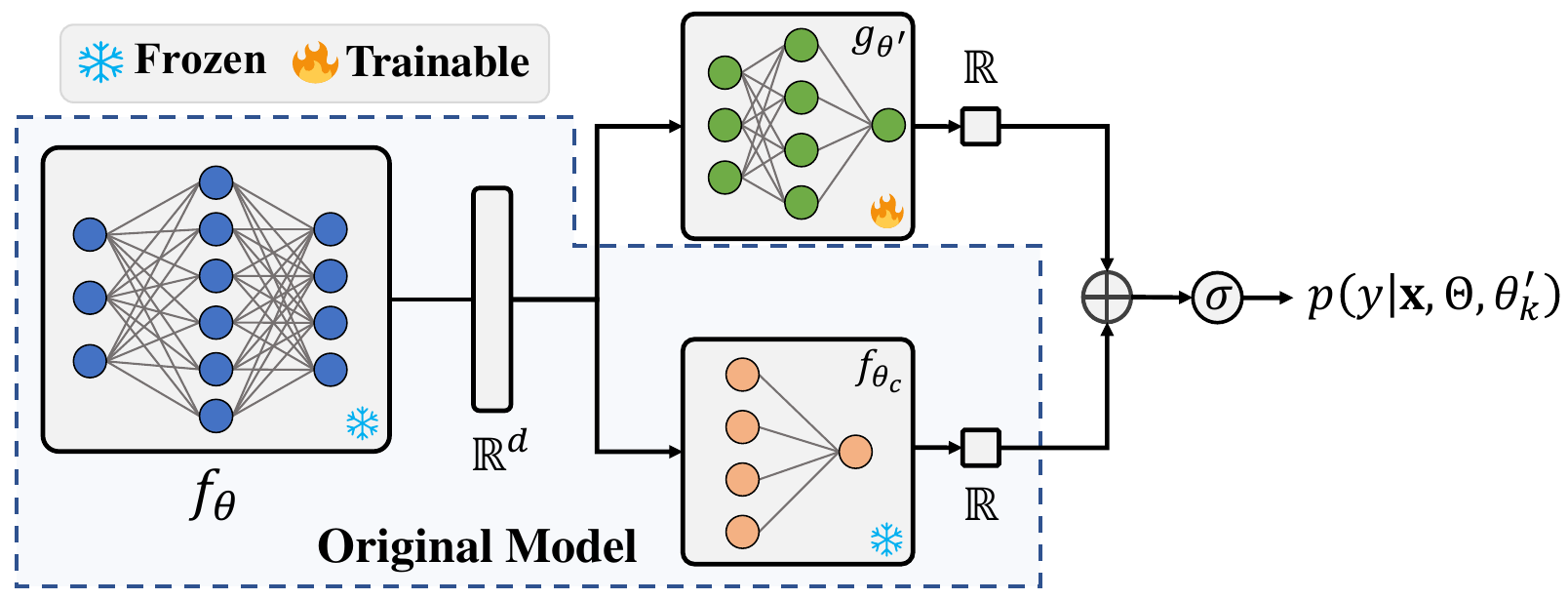}
  \vspace{-0.6cm}
  \caption{
    The architecture of the appended model. $\sigma$ means the sigmoid layer.
  }
  \label{fig:appended_model}
  \vspace{-0.6cm}
\end{figure}
\subsubsection{Post-train Bayesian Strategy}
However, it is not straightforward to attack AIGI detectors in such a Bayesian manner due to several factors. First, the Bayesian posterior for DNNs is a high-dimensional distribution due to a very large number of parameters of DNNs~\cite{izmailov2021bayesian}. Hence computing and sampling the posterior distribution is an intractable problem. Albeit feasible for approximately sampling the posterior via variational inference or Markov Chain Monte Carlo (MCMC), it is computationally slow and expensive in such a high-dimensional space. Furthermore, to improve the accuracy and generalization of the AIGI detectors, there is a growing inclination to train AIGI detectors on large-scale datasets~\cite{zhu2024genimage,he2021forgerynet}. From the perspective of end-users, it is not desirable to re-train a surrogate model on large-scale datasets for attack. 

Therefore, we propose a \textit{post-train} Bayesian strategy to turn a single surrogate into a Bayesian one, without the need for re-training. The parameters over the pre-trained surrogate are represented as $\Theta=[\theta,\theta_c]$, in which $f_{\theta}$ represents the feature extraction backbone, and $f_{\theta_{c}}$ represents the fully-connected layer network for classification. As shown in \cref{fig:appended_model}, we fix the pre-trained surrogate and append a tiny Bayesian component $g_{\theta'}$ behind the feature extraction backbone $f_{\theta}$. The new logits can be computed via a skip connection:
\begin{equation}
    \label{eq:skip}
    \operatorname{logits}=g_{\theta^{\prime}}(f_{\theta}(\mathbf{x}))+f_{\theta_c}(\mathbf{x}).
\end{equation}

We choose to apply Bayesian Model Averaging to optimize the appended Bayesian model:

\begin{align}
\label{eq:post_Bay}
     &\mathbb{E}_{\theta^{\prime} \sim p(\theta^{\prime} \mid \mathcal{D},\Theta)} p\left(y \mid  \mathbf{x}, \Theta, \theta^{\prime} \right) 
     \approx \frac{1}{K} \sum_{k = 1}^{K} p\left( y \mid \mathbf{x}, \Theta, \theta_{k}^{\prime}\right),\nonumber\\
     &\theta_{k}^{\prime} \sim p(\theta^{\prime} \mid \mathcal{D},\Theta),
\end{align}
where $K$ is the number of appended models. $\Theta$ is fixed to avoid re-training. We surprisingly find that adopting a simple MLP layer for appended models works well in all cases, hence training the appended models is much faster than re-training a surrogate. Finally, the frequency-based post-train Bayesian attack can be conducted with iterative gradient-based methods: 
\begin{align}
\label{eq:postBay_attack}
 \Tilde{\mathbf{x}}^{i+1}=\Tilde{\mathbf{x}}^{i} - \alpha \cdot \operatorname{sign}\{\frac{1}{K}  \sum_{k=1}^{K}\nabla \,  log \, p(y \mid \Gamma(\Tilde{\mathbf{x}}^{i}), \Theta, \theta'_k) \}.
\end{align}

\subsubsection{Inference on Bayesian Appended Models}
\label{sec:inference}
$\Theta$ is frozen after pre-training. 
We use Stochastic Gradient Adaptive Hamiltonian Monte Carlo\cite{springenberg_bayesian_2016} to sample appended model $\theta'$ in each iteration:

\begin{align}
\label{eq:SGAHMC}
&\theta'_{t+1} = \theta'_t - \sigma^2\mathbf{C}^{-1/2}_{\theta'_t}\mathbf{h}_{\theta'_t} + \mathbf{N}(0, 2F\sigma^3\mathbf{C}^{-1}_{\theta'_t} - \sigma^4\mathbf{I}), \nonumber\\
&\mathbf{C}_{\theta'_t} \leftarrow (1 - \tau^{-1})\mathbf{C}_{\theta'_t} + \tau^{-1}\mathbf{h}_{\theta'_t}^2,
\end{align}
where $\sigma$ represents the step size, $F$ denotes the friction coefficient, $\mathbf{h}$ is the stochastic gradient of the system, $\mathbf{N}$ represents a Normal distribution, $\mathbf{I}$ stands for an identity matrix, $\mathbf{C}$ is a pre-conditioner updated through an exponential moving average, and $\tau$ is chosen automatically~\cite{springenberg_bayesian_2016}.

\subsection{Hybrid Adversarial Attack}
Despite detecting fake fingerprints in the frequency domain, some works also extract fingerprint features in the spatial domain~\cite{wang2020cnn}. We hence incorporate the attack gradient from the frequency domain with the spatial gradient to further improve the adversarial transferability across different domains. Specifically, we define the hybrid attack as: 
\begin{align}
\label{eq:hybrid}
 \Tilde{\mathbf{x}}^{i+1}&=\Tilde{\mathbf{x}}^{i} - \alpha \cdot \operatorname{sign}\{\frac{1}{K}  \sum_{k=1}^{K}(g^i_k + d^i_k) \},  \\
 \label{eq:g1}
  g^i_k &= \frac{1}{N}\sum_{n=1}^{N} \nabla \,  log \, p(y \mid \Gamma(\Tilde{\mathbf{x}}^{i}_{n-1}), \Theta, \theta'_k), \Tilde{\mathbf{x}}^i_0 = \Tilde{\mathbf{x}}^i, \\
  \label{eq:g2}
  d^i_k &= \nabla \,  log \, p(y \mid \Tilde{\mathbf{x}}^{i}, \Theta, \theta'_k),
\end{align}
where $g^i_k$ and $d^i_k$ are the gradients computed in the frequency domain and spatial domain respectively. For frequency gradient, we conduct random spectrum transformation with $N$ times to get more diverse spectrums. Our proposed method leverages both spatial attack gradients and frequency attack gradients in a Bayesian manner, aiming to further narrow the discrepancy between surrogate models and victim models. The complete algorithm of our method is presented in \cref{alg:FPBA}. An overview illustration of FPBA is shown in \cref{fig:overview}.

\setlength{\textfloatsep}{3pt}
\begin{algorithm}[tb]
\SetAlgoLined
\textbf{Input}: $\mathbf{x}$: training data; $N_{tra}$: the number of training iterations; $M_{\theta'}$: sampling iterations for $\theta'$;$\Theta$: parameters over pre-trained surrogate model; $\{\theta'_1, \dots, \theta'_K\}$: parameters over appended models; $K$: the number of appended models\;
\textbf{Output}: The adversarial example $\Tilde{\mathbf{x}}$\;
\tcp{Post-train Bayesian Optimization}
Randomly initialize $\{\theta'_1, \dots, \theta'_K\}$\;
\For{j = 1 to $N_{tra}$}{
    \For{n = 1 to $K$}{
        Randomly sample a mini-batch data $\{\mathbf{x}, y\}_j$\;
        Compute $\mathbf{h}_{\theta'_k} = \frac{\partial log p(y|\mathbf{x},\Theta,\theta'_k)}{\partial\theta'_k}$\;    
        \For{t = 1 to $M_{\theta'}$}{
        Update $\theta'_k$ with $\mathbf{h}_{\theta'_k}$ via \cref{eq:SGAHMC}\;
        }
    }
}
\Return $\{\theta'_1, \dots, \theta'_K\}$\;
\tcp{Frequency-based Post-train Bayesian Attack}
$\Tilde{\mathbf{x}}^{0} = \mathbf{x}$ \;
\For{i = 1 to $I$}{
    \For{n = 1 to $N$}{
        Get spectrum transformation output $\Gamma(\Tilde{\mathbf{x}}^{i})$ using \cref{eq:st} \;              
}
    \For{k = 1 to $K$}{
    Average frequency gradient $g_k$ using \cref{eq:g1} \;
    Calculate spatial gradient $d_k$ using \cref{eq:g2} \;
    }
    Sample $\Tilde{\mathbf{x}}^{i+1}$ from $\Tilde{\mathbf{x}}^{i}$ via \cref{eq:hybrid} \;
}
\Return $\Tilde{\mathbf{x}}$\;
\caption{Inference on FPBA}
\label{alg:FPBA}
\end{algorithm}

\section{Experimets}
\label{sec:exp}
\subsection{Experimental Settings}
\noindent{\textbf{Datasets: }}
We chose three generated image datasets created by a wide range of generative models. Synthetic LSUN is a commonly used dataset proposed by CNNSpot~\cite{wang2020cnn}, containing 360k real images from LSUN and 360k fake images generated by ProGAN~\cite{progan}. GenImage~\cite{zhu2024genimage} is a recently proposed large-scale dataset, containing 1331k real images and 1350k fake images generated by eight generative models. Following the protocol in \cite{zhu2024genimage}, we employ a subset of GenImage, collecting 162k real images from Imagenet~\cite{deng2009imagenet} and 162k Stable Diffusion(SD) V1.4~\cite{sd} generated images for training. The images generated by the other generators are used for testing in \cref{sec:cross}. To verify our proposed attack is a universal threat across AIGI and Deepfake detection, we also employ the synthetic FFHQ face dataset proposed by~\cite{shamshad2023evading}. The Deepfake dataset consists of 50k real face images from FFHQ~\cite{ffhq} and 50k generated face images generated by StyleGAN2~\cite{ffhq}. After training, we collect only the correctly classified testing samples for attack in evaluation.

\begin{table}[th]
\caption{Datasets employed for training each detector.}
\label{tab:dataset}
\begin{tabular}{cccc}
\hline
             & Synthetic LSUN             & GenImage                   & Synthetic FFHQ             \\ \hline
CNNSpot      & \checkmark & \checkmark & \checkmark \\
DenseNet     & \checkmark & \checkmark & \checkmark \\
EfficientNet & \checkmark & \checkmark & \checkmark \\
MobileNet    & \checkmark & \checkmark & \checkmark \\
Spec         & \checkmark & \checkmark & \checkmark \\
DCTA         & \checkmark & \checkmark & \checkmark \\
ViT          & \checkmark & \checkmark & \checkmark \\
Swin         & \checkmark & \checkmark & \checkmark \\
GramNet      & \checkmark & - &   -    \\
LGrad        & \checkmark & - &  -  \\
LNP          & \checkmark & - & -  \\
UnivFD       & \checkmark & - & - \\
DNF          & \checkmark & - & - \\
FreqNet      & \checkmark & - & - \\
FreqMask     & \checkmark & - & - \\
DIRE         & - & \checkmark & - \\
\hline                             
\end{tabular}
\end{table}

\noindent{\textbf{Evaluated Models:}} 
We extensively evaluate the transferability of adversarial examples by 17 state-of-the-art AIGI detectors, including heterogeneous model architectures and various detection methods. For evaluating on different model architectures, we choose CNN-based detectors CNNSpot~\cite{wang2020cnn}, MobileNet~\cite{howard2017mobilenets}, EfficientNet~\cite{tan2019efficientnet} and DenseNet~\cite{huang2017densely}, and ViT-based detectors Vision Transformer(ViT)~\cite{vit} and Swin-Transformer(Swin-ViT)~\cite{liu2021swin}. For evaluating on various detection methods, we use frequency-based detectors DCTA~\cite{frank2020leveraging}, Spec~\cite{zhang2019detecting}, \revision{FreqNet~\cite{tan2024frequency}, and FreqMask~\cite{doloriel2024frequency}}, gradient-based detectors LGrad~\cite{tan2023learning}, CLIP-based detectors UnivFD~\cite{ojha2023towards}, and diffusion-based detectors DNF~\cite{zhang2023diffusion}, \revision{DIRE~\cite{wang2023dire} and AEROBLADE~\cite{ricker2024aeroblade}}. In addition, LNP~\cite{liu2022detecting} extracts the noise pattern of images and GramNet~\cite{GramNet} learns the global texture representation. We also consider them as victim models. \revision{The datasets employed for training each detector are summarized in \cref{tab:dataset}. Note that AEROBLADE is a training-free method and not trained on the listed datasets.}

\noindent{\textbf{Compared Methods: }} \revisionb{We adopt gradient-based methods, I-FGSM~\cite{I-FGSM}, PGD~\cite{pgd}, MI~\cite{dong2018boosting} and frequency-based methods S$^2$I~\cite{long2022frequency}, SSAH~\cite{luo2022frequency} and ensemble-based methods ENS~\cite{dong2018boosting} and SVRE~\cite{svre}. Because attacks against Deepfake detectors are the most similar adversaries to ours, we also consider state-of-the-art Deepfake detection attacks \revision{FaceAttacker~\cite{jia2022exploring}}, Fakepolisher~\cite{huang2020fakepolisher} and TraceEvader~\cite{wu2024traceevader} as baselines. Fakepolisher and TraceEvader remove the Deepfake traces and we follow their default settings.} For other iterative attacks, we run 10 iterations with step size $\alpha=2/255$ under $l_{\infty}$ perturbation budget of $8/255$ for all these attacks. 

\noindent{\textbf{Implementation Details:}} For MobileNet, EfficientNet, DenseNet, ViT and Swin-Transformer(Swin-ViT), we train them following the default setting in CNNSpot~\cite{wang2020cnn}. Specifically, we choose classifiers pre-trained on ImageNet, and train them with Adam optimizer using Binary Cross-Entropy loss with an initial learning rate of 0.0001. \revisionb{For a fair evaluation, we follow the same data augmentation strategy used in CNNSpot~\cite{wang2020cnn} to improve the models' generalization and robustness.} Before cropping, images are blurred with $\sigma \sim \text{Uniform}[0, 3]$ with 10\% probability, and JPEG-ed with 10\% probability. We transformed images to 224 pixels on the Synthetic LSUN(ProGAN) and GenImage(SD) datasets following CNNSpot~\cite{wang2020cnn}. On the FFHQ (StyleGAN2) dataset, we resized images to 224 pixels to ensure the integrity of the real/fake faces. Subsequently, we apply ImageNet normalization across three datasets. For other AIGI detectors, we use the pre-trained model from their official code. For post-train Bayesian optimization, we follow the default setting in~\cite{bbc}. Although BNNs theoretically necessitate sampling numerous for inference, in practice, we find the number of models $K=3$ is adequate. Opting for a larger number of appended models escalates computational overhead; thus, we opt for $K=3$. For frequency-based attack, we set the tuning factor $\rho$ = 0.5 for $\mathcal{M}$, the standard deviation $\sigma$ of $\xi$ is set to the value of $\epsilon$, following \cite{long2022frequency}. All experiments were conducted on 4 NVIDIA GeForce RTX 3090s. 

\begin{table*}[!htb]

\caption{The attack success rate(\%) on CNN-based, ViT-based and Frequency-based models on the Synthetic LSUN and GenImage subset. “Average” was calculated as the average transfer success rate over all victim models except for the surrogate model. We mark the white-box attack results in gray, and black-box attack results are not marked with colors.}
\centering
\label{tab:commands}
      \setlength{\tabcolsep}{3pt}
    \renewcommand\arraystretch{0.95}	
    \scalebox{1}{%
  \begin{tabular}{c|c|ccccccccc|c}
   \hline
 & \textbf{Surrogate Model} & \multicolumn{1}{c|}{\textbf{Attack Methods}} & \textbf{CNNSpot} & \textbf{DenseNet} & \textbf{EfficientNet} & \textbf{MobileNet} & \textbf{Spec} & \textbf{DCTA} & \textbf{ViT} & \textbf{Swin} & \textbf{Average}\\ 
     \hline
\multicolumn{1}{c|}{} & \multicolumn{1}{c|}{} & \multicolumn{1}{c|}{IFGSM} & \colorbox{mygray}{52.1} & 51.3 & 24.6 & 48.4 & 17.7 & 48.0 & 35.2 & 46.2 & 38.7 \\
\multicolumn{1}{c|}{} & \multicolumn{1}{c|}{} & \multicolumn{1}{c|}{MIFGSM} & \colorbox{mygray}{52.1} & 51.5 & 27.9 & 47.6 & \textbf{26.1} & \textbf{49.8} & 37.5 & 42.7 & 40.4 \\
\multicolumn{1}{c|}{} & \multicolumn{1}{c|}{} & \multicolumn{1}{c|}{PGD} & \colorbox{mygray}{78.3} & 77.6 & 49.4 & 73.0 & 25.6 & 47.6 & 41.0 & 70.2 & 54.9 \\
\multicolumn{1}{c|}{} & \multicolumn{1}{c|}{CNNSpot} & \multicolumn{1}{c|}{S$^2$I} & \colorbox{mygray}{97.8} & 86.4 & 61.5 & 78.6 & 20.5 & 40.8 & 11.7 & 74.5 & 53.4 \\

\multicolumn{1}{c|}{} & \multicolumn{1}{c|}{} & \multicolumn{1}{c|}{\revisionb{SSAH}} & \colorbox{mygray}{\revisionb{97.8}} & \revisionb{2.5} & \revisionb{1.3} & \revisionb{3.0} & \revisionb{0.8} & \revisionb{1.0} & \revisionb{1.0} & \revisionb{0.7} & \revisionb{1.4} \\  

\multicolumn{1}{c|}{} & \multicolumn{1}{c|}{} & \multicolumn{1}{c|}{\textbf{FPBA(Ours)}} & \colorbox{mygray}{\textbf{98.9}} & \textbf{98.0} & \textbf{76.4} & \textbf{95.9} & 19.8 & 48.0 & \textbf{51.5} & \textbf{94.8} & \textbf{69.2}\\  \cline{2-12}

\multicolumn{1}{c|}{} & \multicolumn{1}{c|}{} & \multicolumn{1}{c|}{IFGSM} & 14.3 & 17.8 & 22.4 & \colorbox{mygray}{75.4} & 17.2 & 34.9 & 9.2 & 18.4 & 19.2 \\
\multicolumn{1}{c|}{} & \multicolumn{1}{c|}{} & \multicolumn{1}{c|}{MIFGSM} & 23.8 & 23.9 & 28.6 & \colorbox{mygray}{75.4} & 19.6 & \textbf{42.0} & \textbf{15.5} & 24.8 & 25.5 \\
\multicolumn{1}{c|}{} & \multicolumn{1}{c|}{} & \multicolumn{1}{c|}{PGD} & 20.3 & 26.0 & 20.6 & \colorbox{mygray}{97.6} & \textbf{25.0} & 40.4 & 9.7 & 29.9 & 24.6 \\
\multicolumn{1}{c|}{} & \multicolumn{1}{c|}{MobileNet} & \multicolumn{1}{c|}{S$^2$I} & 14.4 & 17.5 & 34.7 & \colorbox{mygray}{97.8} & 16.3 & 25.4 & 4.3 & 14.6 & 18.2 \\

\multicolumn{1}{c|}{} & \multicolumn{1}{c|}{} & \multicolumn{1}{c|}{\revisionb{SSAH}} & \revisionb{0.6} & \revisionb{1.0} & \revisionb{1.5} & \colorbox{mygray}{\revisionb{99.2}} & \revisionb{0.5} & \revisionb{0.8} & \revisionb{1.0} & \revisionb{1.1} & \revisionb{0.9}\\

\multicolumn{1}{c|}{\multirow{-12}{*}{\rotatebox{90}{ Synthetic LSUN (ProGAN)}}} & \multicolumn{1}{c|}{} & \multicolumn{1}{c|}{\textbf{FPBA(Ours)}}   & \textbf{32.2} & \textbf{40.4} & \textbf{51.7} & \colorbox{mygray}{\textbf{99.6}} & 22.6 & 36.4 & 12.3 & \textbf{44.6} & \textbf{34.3} \\  \cline{1-12}

\multicolumn{1}{c|}{} & \multicolumn{1}{c|}{} & \multicolumn{1}{c|}{IFGSM} & \colorbox{mygray}{78.3} & 71.5 & 30.9 & 34.0 & 27.1 & 28.3 & 11.0 & 27.8 & 32.9 \\
\multicolumn{1}{c|}{} & \multicolumn{1}{c|}{} & \multicolumn{1}{c|}{MIFGSM} & \colorbox{mygray}{78.3} & 75.3 & 30.1 & 35.8 & 27.8 & 28.3 & 15.0 & 28.2 & 34.4 \\
\multicolumn{1}{c|}{} & \multicolumn{1}{c|}{} & \multicolumn{1}{c|}{PGD} & \colorbox{mygray}{99.3} & 88.0 & 48.5 & 52.2 & \textbf{48.6} & 49.9 & 15.4 & 48.1 & 50.1 \\
\multicolumn{1}{c|}{} & \multicolumn{1}{c|}{CNNSpot} & \multicolumn{1}{c|}{S$^2$I} & \colorbox{mygray}{98.0} & 91.0 & \textbf{66.8} & 64.7 & 36.7 & \textbf{50.1} & 16.9 & \textbf{53.0} & 54.2\\

\multicolumn{1}{c|}{} & \multicolumn{1}{c|}{} & \multicolumn{1}{c|}{\revisionb{SSAH}} & \colorbox{mygray}{\revisionb{94.8}} & \revisionb{5.4} & \revisionb{0.5} & \revisionb{1.1} & \revisionb{0.6} & \revisionb{3.6} & \revisionb{0.7} & \revisionb{0.3} & \revisionb{1.7}\\

\multicolumn{1}{c|}{} & \multicolumn{1}{c|}{} & \multicolumn{1}{c|}{\textbf{FPBA(Ours)}} & \colorbox{mygray}{\textbf{100}} & \textbf{96.6} & 59.4 & \textbf{67.3} & 47.9 & 49.8 & \textbf{21.1} & 51.2 & \textbf{56.2}\\  \cline{2-12}

\multicolumn{1}{c|}{} & \multicolumn{1}{c|}{} & \multicolumn{1}{c|}{IFGSM} & 9.0 & 8.9 & 8.9 & \colorbox{mygray}{58.9} & 9.1 & 9.1 & 4.4 & 8.5 & 8.3\\
\multicolumn{1}{c|}{} & \multicolumn{1}{c|}{} & \multicolumn{1}{c|}{MIFGSM} & 9.7 & 9.2 & 8.9 & \colorbox{mygray}{59.0} & 9.5 & 9.0 & 5.9 & 8.8 & 8.7\\
\multicolumn{1}{c|}{} & \multicolumn{1}{c|}{} & \multicolumn{1}{c|}{PGD} & 42.3 & 39.9 & 36.4 & \colorbox{mygray}{97.1} & \textbf{48.4} & \textbf{49.8} & 14.2 & 40.0 & 38.7\\
\multicolumn{1}{c|}{} & \multicolumn{1}{c|}{MobileNet} & \multicolumn{1}{c|}{S$^2$I} & 35.6 & 33.1 & 26.8 & \colorbox{mygray}{78.7} & 24.1 & 43.4 & 6.2 & 28.0 & 28.2 \\

\multicolumn{1}{c|}{} & \multicolumn{1}{c|}{} & \multicolumn{1}{c|}{\revisionb{SSAH}} & \revisionb{1.1} & \revisionb{2.0} & \revisionb{0.6} & \colorbox{mygray}{\revisionb{96.7}} & \revisionb{0.5} & \revisionb{1.9} & \revisionb{1.2} & \revisionb{2.2} & \revisionb{1.4}\\

\multicolumn{1}{c|}{\multirow{-12}{*}{\rotatebox{90}{GenImage(SD)}}} & \multicolumn{1}{c|}{} & \multicolumn{1}{c|}{\textbf{FPBA(Ours)}} & \textbf{49.7} & \textbf{46.8} & \textbf{44.0} & \colorbox{mygray}{\textbf{100}} & 44.1 & 49.4 & \textbf{16.6} & \textbf{45.0} & \textbf{42.2} \\  \cline{1-12}

\hline
\end{tabular}}
\vspace{-0.5cm}
\end{table*}

\subsection{Attack on Spatial-based and Frequency-based Detectors}
We report the attack performance against spatial-based and frequency-based detectors in \cref{tab:commands}. Under the white-box setting, our proposed method FPBA achieves the highest attack success rate in all cases, and outperforms other competitive methods. Specifically, FPBA gets an average white-box success rate across different datasets and models as high as 99.6\%, while S$^2$I, PGD, IFGSM and MIFGSM only have 93.1\%, 93.1\%, 66.2\% and 66.2\%. \revisionb{Although SSAH has a relatively high white-box success rate (97.7\%), it fails to transfer the adversarial examples to black-box models. Under the black-box setting, FPBA still achieves the highest average transfer success rate of 50.5\%, surpassing the IFGSM, MIFGSM, PGD, S$^2$I and SSAH by a big margin of 25.7\%, 23.3\%, 8.4\%, 11.9\% and 49.1\%.} Note that our proposed FPBA significantly outperforms the SOTA frequency-based attack S$^2$I and SSAH, demonstrating that FPBA is a stronger frequency-based attack against AIGI detectors.

\begin{table}[!tb]
\caption{\revisionb{The attack success rate(\%) compared with attacks against Deepfake detectors(only fake ASR).DenNet, EffNet, MobNet refers to DenseNet, EfficientNet, MobileNet respectively.}} 
\vspace{-0.6cm}
\small
\label{tab:deepfake_attack}
\begin{center}
\resizebox{1\linewidth}{!}{
    \setlength{\tabcolsep}{3pt}
    \renewcommand\arraystretch{1}	
    \scalebox{1}{%
\begin{tabular}{c|c|cccccc|c}
\hline
\textbf{Datasets} & \textbf{Attack} &  \textbf{CNNSpot} & \textbf{DenNet} & \textbf{EffNet} & \textbf{MobNet}   & \textbf{ViT}  & \textbf{Swin}  &\textbf{Ave.}\\
\hline
\multicolumn{1}{c|}{} & FakePolisher & 94.5 & 98.3 & 71.6 & 88.6 & 15.1 & 99.5 & 77.9  \\
\multicolumn{1}{c|}{LSUN} & TraceEvader & 0.6 & 1.3 & 4.5 & 4.8 & 8.10 & 11.4 & 1.9\\

\multicolumn{1}{c|}{(ProGAN)} & \revision{FaceAttacker} & \revision{52.9} & \revision{30.9} & \revision{3.7} & \revision{8.8} & \revision{2.8} & \revision{18.9} & \revision{19.7}\\

\multicolumn{1}{c|}{} & FPBA & \textbf{100.0} & \textbf{100.0} & \textbf{98.2} & \textbf{99.5} & \textbf{38.6} &\textbf{99.9} & \textbf{89.4}\\ \cline{1-9}
\multicolumn{1}{c|}{} & FakePolisher & 93.5 & 93.2 & 12.2 & 95.5 & 33.5 & 97.6 & 70.9  \\
\multicolumn{1}{c|}{GenImage} & TraceEvader & 46.1 & 34.3 & 9.4 & 37.0 & 15.2 & 44.3 & 31.1\\

\multicolumn{1}{c|}{(SD)} & \revision{FaceAttacker} & \revision{94.3} & \revision{79.8} & \revision{57.9} & \revision{77.0} & \revision{12.4} & \revision{61.2} & \revision{63.7}\\

\multicolumn{1}{c|}{} & FPBA & \textbf{100.0} & \textbf{99.7} & \textbf{97.7} & \textbf{98.9} & \textbf{42.1} &\textbf{97.9} & \textbf{89.4}\\
\hline
\end{tabular}}}
\end{center}
\vspace{-0.2cm}
\end{table}

\begin{figure}[!t]
  \centering
  \includegraphics[width=1\linewidth]{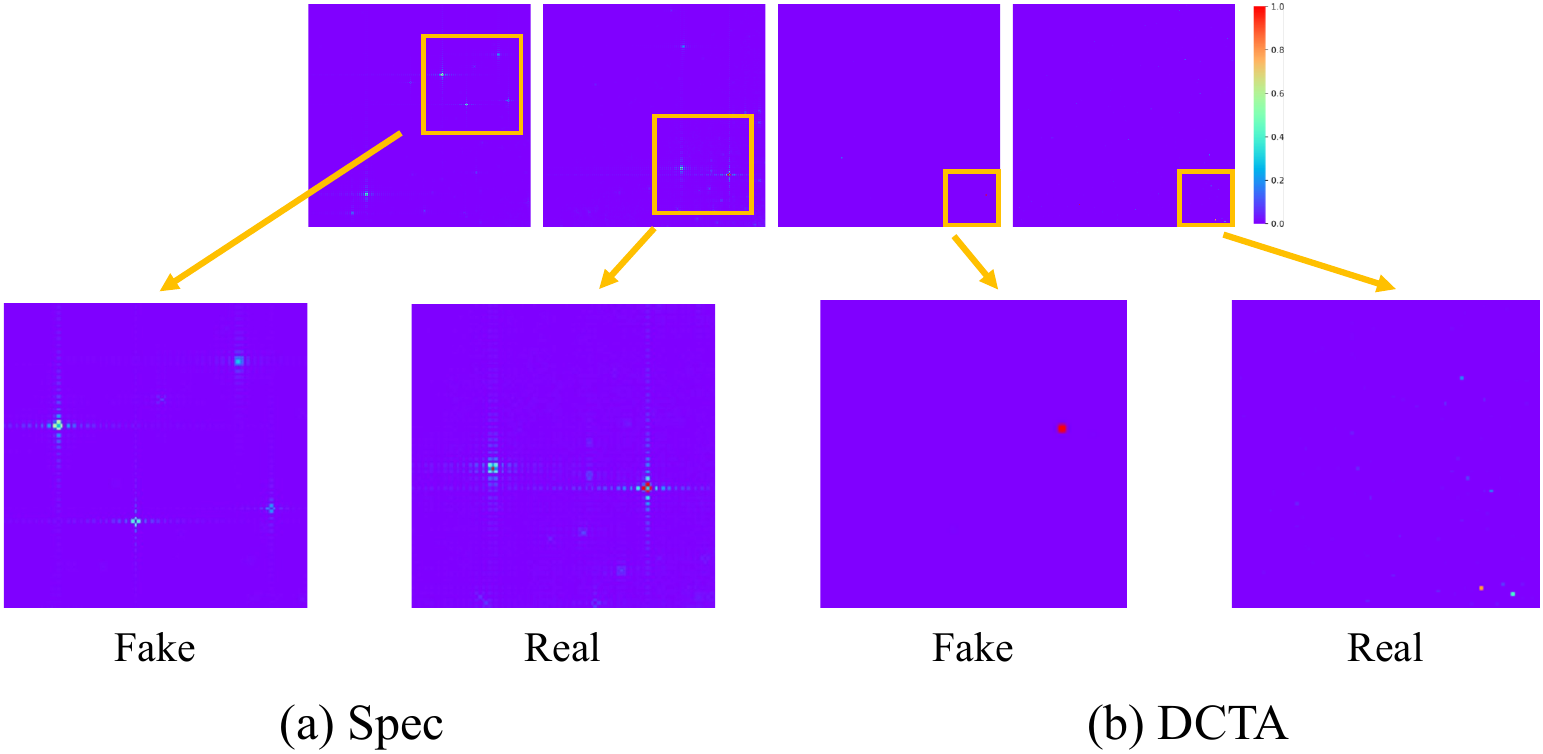}
  \vspace{-0.6cm}
  \caption{
Visualization of the sensitive frequency components of real and fake images (average 1000 images on LSUN/ProGAN datasets) for frequency-based models. The frequency components of frequency-based models are highly sparse in comparison with spatial-based models. The color value represents the absolute gradient value of the model loss function after max-min normalization.}
  \label{fig:fre_ssm}
\end{figure}

\subsubsection{\revisionb{Comparison with Attacks against Deepfake Detectors}} 
\begin{table}[tbh]
\caption{The attack success rate(\%) compared with ensemble attack methods on Synthetic LSUN datasets. $^1$ means setting (1), i.e., taking an ensemble of CNNSpot, MobileNet, and EfficientNet; $^2$ means setting (2), i.e., taking an ensemble of CNNSpot, MobileNet and ViT; $^3$ means setting (3), i.e., taking an ensemble of CNNSpot, DCTA and ViT. $^4$ means only using CNNSpot as a surrogate model. 
} 
\label{tab:exper_ensemble}
\vspace{-0.4cm}
\begin{center}
    \setlength{\tabcolsep}{2.1pt}
    \renewcommand\arraystretch{0.99}	
    \scalebox{0.95}{%
\begin{tabular}{c|cccccccc|c}
\hline
 \textbf{Attack}& \textbf{CNNSpot} & \textbf{DenNet} & \textbf{EffNet} & \textbf{MobNet} & \textbf{Spec}  & \textbf{DCTA} & \textbf{ViT} & \textbf{Swin} &\textbf{Ave}\\
\hline
ENS$^1$  & 55.9 & 45.4 & 69.9 & 76.2 & 16.8 & 42.9 & 23.3 & 43.7 &34.4\\
SVRE$^1$ & 63.7 & 64.5 & 75.5 & 72.9 & 10.3 & 30.4 & 28.6 & 64.2 &39.6\\
ENS$^2$  & 57.3 & 58.6 & 54.1 & 92.8 & 19.5 & 38.5 & \textbf{77.9} & 59.7 &46.0\\
SVRE$^2$ & 77.7 & 77.9 & \textbf{76.4} & \textbf{97.8} & 13.9 & 33.6 & 70.1 & 78.1 &56.0\\
ENS$^3$  & 9.7  & 6.3  & 4.8  & 10.1 & \textbf{37.0} & 72.9 & 20.6 & 5.5 &12.7\\ 
SVRE$^3$ & 22.6 & 12.9 & 8.1  & 13.2 & 29.4 & \textbf{87.0} & 24.0 & 9.6 &14.6\\
\textbf{FPBA$^4$} & \textbf{98.9}  &  \textbf{98.0} & \textbf{76.4} & 95.9 & 19.8 &  48.0 & 51.5 & \textbf{94.8} &\textbf{69.2} \\
\hline
\end{tabular}}
\end{center}
\vspace{0.0cm}
\end{table}

\begin{table*}[!tbh]
\caption{\revisionb{Transfer-based attack against SOTA detectors on Synthetic LSUN (ProGAN) datasets. The surrogate model is chosen as CNNSpot~\cite{wang2020cnn}.}} 
\vspace{-0.5cm}
\label{tab:exper_sota}
\begin{center}
    \setlength{\tabcolsep}{3pt}
    \renewcommand\arraystretch{0.99}	
    \scalebox{1}{%
\begin{tabular}{c|ccccccc|c}
\hline
\textbf{Attack} & \textbf{GramNet} & \textbf{LGrad} & \textbf{LNP} & \textbf{UnivFD} & \textcolor{black}{\textbf{DNF}} & \revision{FreqNet} & \revision{FreqMask} & \revision{\textbf{Ave}} \\
\hline
MIFGSM  & 9.7 & 6.6 & 38.5 & 24.2 & \textcolor{black}{46.7}  & \revision{10.5} & \revision{50.3}  & \revision{26.6} \\
 PGD     & 67.7 & 50.2 & \textbf{47.4} & 11.5 & \textcolor{black}{48.3} & \revision{\textbf{48.9}} & \revision{77.2}  & \revision{50.2} \\
S$^2$I  & 49.1 & 49.1 & 28.3 & 8.3 & \textcolor{black}{43.0} & \revision{45.3} & \revision{84.2} & \revision{43.9} \\ 
\textbf{FPBA} & \textbf{87.8} & \textbf{50.2} & 39.5 & \textbf{16.6} & 
\textbf{\textcolor{black}{49.8}} & \revision{47.1} & \revision{\textbf{98.4}}   & \revision{\textbf{55.6}} \\
\hline
\end{tabular}}
\end{center}
\vspace{-0.6cm}
\end{table*}

\begin{table}[!tbh]
\caption{Benign accuracy of models trained on SD V1.4 and evaluated on different generated data. `Real/Fake' means the accuracy(\%) on evaluating real/fake testing data.} 
\label{tab:exper_unseenDomainAcc}
\vspace{-0.5cm}
\begin{center}
    \setlength{\tabcolsep}{2pt}
    \renewcommand\arraystretch{0.99}	
    \scalebox{0.99}{%
\begin{tabular}{c|c|cccccc|c}
\hline
\textbf{Detector} & \textbf{Acc} & \textbf{Midj.} & \textbf{SDv1.4} & \textbf{SDv1.5} & \textbf{ADM}  & \textbf{Wukong} & \textbf{BigGAN} & \textbf{Ave}\\
\hline
\multicolumn{1}{c|}{} & All       & 60.5 & 100.0 & 99.9 & 50.8 & 99.7 & 50.0 & 76.8 \\
\multicolumn{1}{c|}{} & Real  & 100.0 & 100.0 & 100.0 & 100.0 & 100.0 & 99.9 & 100.0 \\
\multirow{-3}{*}{Swin-ViT} & Fake  & 21.0 & 100.0 & 99.9 & 1.5 & 99.3 & 0.1 & 53.6 \\ \cline{1-9}
\multicolumn{1}{c|}{} & All       & 62.0 & 99.6 & 99.5 & 50.3 & 97.9 & 49.8 & 76.5 \\
\multicolumn{1}{c|}{} & Real  & 99.4 & 99.6 & 99.6 & 99.5 & 99.6 & 99.4 & 99.5 \\
\multirow{-3}{*}{CNNSpot} & Fake  & 24.7 & 99.6 & 99.4 & 1.2 & 96.1 & 0.2 & 53.5 \\ \cline{1-9}
\multicolumn{1}{c|}{} & All       & 58.4 & 99.5 & 99.3 & 50.2 & 96.0 & 50.2 & 75.6 \\
\multicolumn{1}{c|}{} & Real  & 99.5 & 99.6 & 99.5 & 99.5 & 99.5 & 99.3 & 99.5 \\
\multirow{-3}{*}{ViT} & Fake  & 17.4 & 99.3 & 99.1 & 0.9 & 92.4 & 1.1 & 51.7 \\
\hline
\end{tabular}
}
\vspace{-0.3cm}
\end{center}
\end{table}
Attacks against Deepfake detectors are similar to our attacks. However, Deepfakes mainly manipulate or synthesize faces, often crafted by GANs. In contrast, AIGI detectors identify any type of visual content generated by diverse techniques, such as GANs, diffusion models, and autoregressive models. The content includes humans, non-human entities, abstract art, landscapes, and more. These differences make it challenging to transfer attacks from DeepFake detectors to AIGI detectors. \revision{For example, techniques like searching for adversarial points on the face manifold~\cite{li2021exploring}, removing facial forgery traces~\cite{huang2020fakepolisher,wu2024traceevader} or simply updating perturbations in the frequency domain to improve attack imperceptibility~\cite{jia2022exploring} may not be applicable for AIGI detectors. To demonstrate this, we compare with FakePolisher~\cite{huang2020fakepolisher}, TraceEvader~\cite{wu2024traceevader} and FaceAttacker~\cite{jia2022exploring}, which are the SOTA attacks for Deepfake detection. Because FakePolisher and TraceEvader are designed for removing forgery traces, we follow their default setting to only report the attack success rate on fake images. As shown in \cref{tab:deepfake_attack}, FPBA consistently achieves the highest attack success rates across all detectors and datasets, outperforming prior Deepfake-based attacks by margins ranging from 11.5\% to 87.5\%.} FakePolisher performs better than TraceEvader and FaceAttack (albeit still less effective than FPBA), but the former largely reduces the image quality, which is very visible and raises suspicion, as shown in \cref{tab:image_quality} and \cref{fig:vis}. This is likely because FakePolisher is primarily trained on the face manifold, limiting its ability to reconstruct fine-grained, identity-preserving details in AIGI images.

\subsubsection{Comparison with Ensemble-based Attacks}
\revisionb{Considering that FPBA attacks an ensemble of appended models, we thus compare it with state-of-the-art ensemble-based methods ENS~~\cite{dong2018boosting} and SVRE~\cite{svre}, i.e., utilizing an ensemble of surrogate models to generate adversarial examples. To examine the impact of different surrogate combinations(CNN-based vs. Vit-based vs. Frequency-based detectors), we conduct an ablation study for ENSEMBLE and SVRE to investigate their impacts in 3 settings,} including (1) taking CNNs as ensemble surrogates (CNNSpot, MobileNet, EfficientNet); (2) taking CNNs and ViTs as ensemble surrogates (CNNSpot, MobileNet, ViT); (3) taking CNNs, ViTs and frequency-based detectors as ensemble surrogates (CNNSpot, DCTA, ViT). Although we can also use more than one architecture for our method, we only use CNNSpot as the surrogate architecture to verify the universal transferability across heterogeneous models. We report the results in \cref{tab:exper_ensemble}. First, compared with setting (1), the heterogeneous model ensemble in setting (2) enhances the transferability of ENSEMBLE and SVRE across ViT-based and CNN-based models, while a more comprehensive ensemble in setting (3) decreases the transferable performance except for the frequency-based detectors. We speculate that there are significant differences in the classification boundaries between frequency-based and spatial-based detectors, thus averaging the ensemble outputs might reduce the original attack strength crafted by a single surrogate. Second, FPBA solely on CNNSpot still achieves competitive results in comparison with ensemble methods. Somewhat surprisingly, the black-box results on FPBA even outperform white-box results on ENSEMBLE and SVRE in some cases. For instance, the adversarial examples only generated by CNNSpot with FPBA get a success rate of 76.4\% on EfficientNet, which is higher than ENSEMBLE$^1$ and SVRE$^1$ generated by an ensemble of CNNSpot, EfficientNet and MobileNet. This demonstrates our proposed method can approximate the true posterior distribution, in which different victim models can be sampled from the posterior.

\subsubsection{Evaluation on Frequency-based Detectors}
Although FPBA has the best average success rate, we find that the attack results are usually not the best on frequency-based models. To further investigate the reason, we plot the spectrum saliency map of frequency-based detectors using \cref{eq:ssm} in \cref{fig:fre_ssm}. Unlike spatial-based detectors, which rely on numerous frequency components to make decisions (see \cref{fig:vis_ssm}), the frequency components focused on in frequency-based models are very sparse. Therefore, the gradient information available from the spatial domain is much richer compared to the frequency domain. Nevertheless, FPBA still gets a top-2 average transfer success rate of 43.1\% against frequency-based detectors, which is only slightly lower 1.5\% than PGD, and much higher 3.3\%, 15.8\% and 17.6\% than S$^2$I, MIFGSM and IFGSM, respectively.

\begin{table}[t]
\caption{The attack success rate (ASR) of cross-generator image detection on different generated subsets. The surrogate model is chosen as Swin-ViT.} 
\label{tab:exper_unseenDomainWhiteAttack}
\vspace{-0.5cm}
\begin{center}
    \setlength{\tabcolsep}{3pt}
    \renewcommand\arraystretch{0.99}	
    \scalebox{0.99}{%
\begin{tabular}{c|cccccc|c}
\hline
\textbf{Victim} & \textbf{Midj.} & \textbf{SDv1.4} & \textbf{SDv1.5} & \textbf{ADM}  & \textbf{Wukong} & \textbf{BigGAN} & \textbf{Ave}\\
\hline
Swin-ViT  & 97.6 & 97.4 & 97.7 & 96.2 & 96.9 & 97.1 & 97.1 \\ 
CNNSpot  & 22.9 & 44.2 & 44.8 & 5.0 & 44.9 & 6.2 & 28.0 \\ 
ViT  & 19.9 & 14.5 & 16.6 & 3.3 & 20.1 & 2.1 & 12.7 \\ 
\hline
\end{tabular}}
\end{center}
\vspace{0.1cm}
\end{table}

\begin{table}[t]
\caption{The attack success rate (ASR) is reported on real images and fake images respectively. The surrogate model is chosen as Swin-ViT.} 
\label{tab:exper_unseenDomainBlackAttack}
\vspace{-0.5cm}
\begin{center}
    \setlength{\tabcolsep}{3pt}
    \renewcommand\arraystretch{0.99}	
    \scalebox{0.92}{%
\begin{tabular}{c|c|ccccccc}
\hline
\multicolumn{1}{c|}{\textbf{Victim}} & \textbf{ASR} & \textbf{Midj.}& \textbf{SDv1.4} & \textbf{SDv1.5} & \textbf{ADM}  & \textbf{Wukong} & \textbf{BigGAN} & \textbf{Ave}\\
\hline
\multicolumn{1}{c|}{} & Real  & 94.9 & 96.4 & 96.2 & 96.0 & 96.0 & 97.0 & 95.9 \\ 
\multicolumn{1}{c|}{\multirow{-2}{*}{Swin-ViT}} & Fake  & 96.5 & 98.3 & 98.7 & 100.0 & 97.9 & 100.0 & 98.3 \\ \cline{1-9}
\multicolumn{1}{c|}{} & Real & 3.0 & 3.4 & 2.9 & 2.0 & 3.6 & 4.6& 3.2 \\ 
\multicolumn{1}{c|}{\multirow{-2}{*}{CNNSpot}} & Fake  & 100.0 & 85.1 & 86.9 & 100.0 & 86.3 & 100.0 & 93.0 \\ \cline{1-9}
\multicolumn{1}{c|}{} & Real & 0.1 & 0.0 & 0.3 & 0.3 & 0.4 & 0.4 & 0.3 \\ 
\multicolumn{1}{c|}{\multirow{-2}{*}{ViT}} & Fake  & 96.1 & 29.0 & 32.9 & 100.0 & 39.8 & 100.0 & 66.3 \\
\hline
\end{tabular}
}
\end{center}
\vspace{-0.3cm}
\end{table}
\subsection{Evaluation on Diverse Detection Strategies}
\subsubsection{\revisionb{Attack Various SOTA Detection Methods}}
Except for detecting in spatial and frequency domains using different backbone models, recent AIGI detectors utilize different latent feature representations to identify synthetic artifacts. Specifically, GramNet~\cite{GramNet} extracts the global texture representation; UnivFD~\cite{ojha2023towards} captures semantic feature using pre-trained CLIP~\cite{CLIP}; LNP~\cite{liu2022detecting} extracts the noise pattern of images; LGrad~\cite{tan2023learning} extract gradient information through a pre-trained model; DNF~\cite{zhang2023diffusion} leverages diffusion noise features, extracting them through an inverse diffusion process~\cite{adm}; \revision{FreqNet~\cite{tan2024frequency} enhances its generalization ability through a lightweight frequency space learning network and FreqMask~\cite{doloriel2024frequency} introduces a frequency-domain mask strategy for data augmentation during training.} These significant differences among various detection methods present challenges for the transferability of adversarial examples. To illustrate this, we examine the adversarial transferability of various detection methods and present the results in \cref{tab:exper_sota}. Overall, the transfer success rate is relatively low compared to the results in \cref{tab:commands}, suggesting that the distinct characteristics of detection methods limit adversarial transferability between each other. Nevertheless, FPBA still achieves the best adversarial transferability.

\begin{table}[]
\caption{Adversarial attack on AEROBLADE and DIRE. Average precision (AP) is used to measure the attack performance. CNNSpot is used as the surrogate model.}
\centering
\label{tab:aerodire}
\begin{tabular}{c|cc}
\hline
\multicolumn{1}{l|}{} & AEROBLADE$\downarrow$ & DIRE$\downarrow$ \\ \hline
Clean & 80.0 & 100.0 \\
MIFGSM                & 55.9    &  58.9   \\
PGD                   & 46.9    &  43.2   \\
SSI                   & 54.6    &  44.5   \\
FPBA                  & \textbf{43.1}    &  \textbf{40.8}   \\ \hline
\end{tabular}
\end{table}

\revision{Very recently, reconstruction error-based methods, such as DIRE~\cite{wang2023dire} and AEROBLADE~\cite{ricker2024aeroblade} have emerged as competitive approaches for detecting diffusion-generated images. Therefore, we adopt them as victim models in our evaluation. As both methods rely on diffusion models, their performance degrades considerably on GAN-generated data. Consequently, we assess their adversarial robustness on the Stable Diffusion subset of the GenImage dataset in \cref{tab:aerodire}. We follow~\cite{ricker2024aeroblade} to report detection performance using average precision (AP). FPBA achieves the lowest average precision (AP) on both AEROBLADE (43.1\%) and DIRE (40.8\%), indicating that FPBA effectively disrupts the reconstruction-error patterns these detectors rely on.}

\subsubsection{Attack Cross-Generator Image Detection}
\label{sec:cross}
One important real-world detection problem is cross-generator image detection, i.e., identifying fake images generated by unseen generative models. We hence evaluate the robustness of cross-generator image detection to investigate whether adversarial examples are a real threat to AIGI detection. We train the detectors on images generated by SD v1.4~\cite{sd} and assess their robustness against adversarial examples, in which the adversarial perturbations are added to the images generated by Midjourney~\cite{Midjourney}, SD V1.4~\cite{sd}, SD V1.5~\cite{sd}, ADM~\cite{adm}, Wukong~\cite{wukong} and BigGAN~\cite{biggan}. Because Swin-ViT achieves the SOTA results on different subsets~\cite{zhu2024genimage}, we use it as the surrogate model.

The benign accuracy and attack performance on unseen source data are reported in~\cref{tab:exper_unseenDomainAcc} and \cref{tab:exper_unseenDomainWhiteAttack} respectively. First, the attack under the white-box setting achieves an almost 100\% success rate. Second, the transfer success rate is positively correlated with the accuracy on unseen source data. The Midjourney, SD v1.4\&v1.5 and Wukong subsets have relatively high accuracy, their corresponding transfer attack success is also relatively high. In contrast, the binary classification accuracy drops to 50\% on ADM and BigGAN subsets, the corresponding adversarial transferability is also limited on them. By looking closely at the accuracy/ASR on real images and fake images(\cref{tab:exper_unseenDomainAcc},\cref{tab:exper_unseenDomainBlackAttack}), we find detectors fail to distinguish fake images on ADM and BigGAN, which means without a good gradient can be followed to misclassify the real image as fake label under the attack. Therefore, we suggest robustness evaluation of cross-generator image detection should be conducted on the test subset with high accuracy as evaluating on the low-accuracy subset is futile.

\begin{figure}[]
  \centering
  \includegraphics[width=1\linewidth]{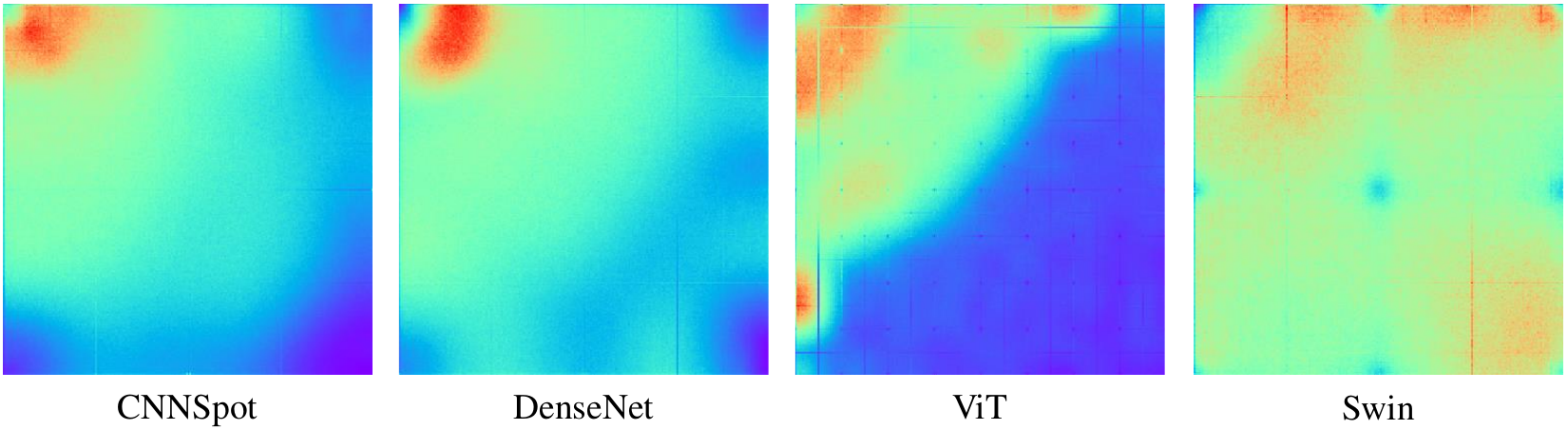}
  \vspace{-0.5cm}
  \caption{\revision{Visualization of the spectrum saliency map (average of adversarial examples generated by FPBA using CNNSpot as the surrogate model) for different AIGI detectors.}}
  \label{fig:visual_transfer}
  \vspace{0.0cm}
\end{figure}

\begin{table}[]
\caption{\revision{Transfer attack success rate on real samples (REAL ASR) and fake samples(FAKE ASR). The surrogate model is chosen as CNNSpot.}}
\centering
\label{tab:sota_analysis}
\begin{tabular}{c|ccccc}
\hline
         & \revision{FreqMask} & \revision{UnivFD} & \revision{GramNet} & \revision{LGrad} & \revision{FreqNet} \\ \hline
\revision{REAL ASR} & \revision{96.7}     & \revision{31.5}   & \revision{76.4}    & \revision{14.6}  & \revision{5.8}     \\
\revision{FAKE ASR} & \revision{100.0}    & \revision{1.7}    & \revision{99.2}    & \revision{85.8}  & \revision{88.4}    \\ 
\revision{ASR}      & \revision{98.4}     & \revision{16.6}   & \revision{87.8}    & \revision{50.2}  & \revision{47.1}    \\ \hline
\end{tabular}
\end{table}

\subsection{\revision{Adversarial Transferability Analysis}}
\revision{As shown in \cref{tab:commands} and \cref{tab:exper_sota}, FPBA exhibit superior adversarial transferability compared to existing attacks. To further investigate the underlying reason, we visualize the spectrum saliency map for different AIGI detectors in \cref{fig:visual_transfer}. The map is computed as the average of adversarial examples generated by FPBA using CNNSpot as the surrogate model. Notably, CNNSpot, DenseNet, and Swin ViT share highly similar frequency-sensitive areas, leading to common vulnerabilities in these domains. In contrast, ViT exhibits fewer overlapping frequency-sensitive features with CNNSpot, resulting in a comparatively lower transfer success rate.}

\revision{Next, considering that the AIGI detection task often suffers from imbalanced recognition accuracy between real and fake samples (as shown in \cref{tab:exper_unseenDomainAcc}), we further investigate the attack success rates of FPBA on real images and fake images separately, as shown in \cref{tab:sota_analysis}. Low-level feature detectors, including FreqMask, FreqNet, GramNet, LGrad, mainly rely on frequency, texture, or spectral cues. We find that these artifacts are more susceptible to adversarial perturbations, resulting in significantly higher attack success rates on fake samples. We suspect that this vulnerability arises from low-level feature detectors overfitting to the limited and monotonous fake patterns, causing the feature space to become highly constrained and low-ranked~\cite{yanorthogonal}. As a result, adversarial perturbations only need to slightly disrupt these brittle and low-dimensional features to deceive the detector, making fake images significantly more vulnerable to attack. In contrast, high-level detectors like UnivFD, which rely on global semantic features, exhibit overall lower transferability. This suggests that models leveraging generalizable semantic representations are inherently more robust. Interestingly, we observe that in UnivFD, fake images are more resistant to adversarial attacks than real ones. We will leave the phenomenon to explore in future work.}

\begin{table}[!tb]
\caption{AT-based defenses against PGD attack.}
\label{tab:at}
\begin{center}
\vspace{-0.1cm}
\begin{tabular}{ccccc}
\hline
Dataset                   & Detectors                                    & AT-methods & Clean & Robust \\\hline
\multirow{4}{*}{ProGAN}   & \multicolumn{1}{c}{\multirow{2}{*}{CNNSpot}} & PGD-AT     & 50.6  & 49.9    \\
                          & \multicolumn{1}{c}{}                         & TRADES     & 50.1  & 6.5    \\
                          & \multirow{2}{*}{UnivFD}                      & PGD-AT     & 69.6  & 0    \\
                          &                                              & TRADES     & 95.5  & 0    \\ \hline
\multirow{4}{*}{GenImage} & \multicolumn{1}{c}{\multirow{2}{*}{CNNSpot}} & PGD-AT     & 50.7  & 49.7    \\
                          & \multicolumn{1}{c}{}                         &  TRADES   & 55.2  & 9.4    \\
                          & \multirow{2}{*}{UnivFD}                      & PGD-AT     & 49.7  & 0.7    \\
                          &                                              & TRADES     & 56.8  & 0   \\ \hline
\end{tabular}
\end{center}
\end{table}

\subsection{\revision{Evaluation against Defense Models}}
\paragraph{\revision{Evaluation against AT-based Methods}}
\revision{Next, we investigate the attack performance against defense models. As adversarial training (AT) is a widely used defense strategy, we first adopt AT for improving the robustness of AIGI detectors. In~\cref{tab:at}, we train two representative detectors—CNNSpot and UnivFD—on the ProGAN and GenImage datasets using two mainstream AT strategies: PGD-AT~\cite{pgd} and TRADES~\cite{trades}. However, we found that the training process of AT fails to converge. Both AT methods cause a substantial drop in clean accuracy, often to around 50\%, which is equivalent to random guessing. Robust accuracy against PGD attacks also remains low, in some cases collapsing to nearly 0\%. This phenomenon can be attributed to the intrinsic data imbalance in AIGI detection: models tend to overfit to low-level synthetic artifacts early in training. When AT is applied, these brittle features collapse quickly, and adversarial examples easily cross the decision boundary for both real and fake classes. As a result, the model fails to learn discriminative and robust representations, producing near-random predictions on clean data and negligible robustness gains.}

\paragraph{Attack Performance under Compression in Real Scenarios}

\begin{figure*}[!htb]
  \centering
  \includegraphics[width=0.75\linewidth]{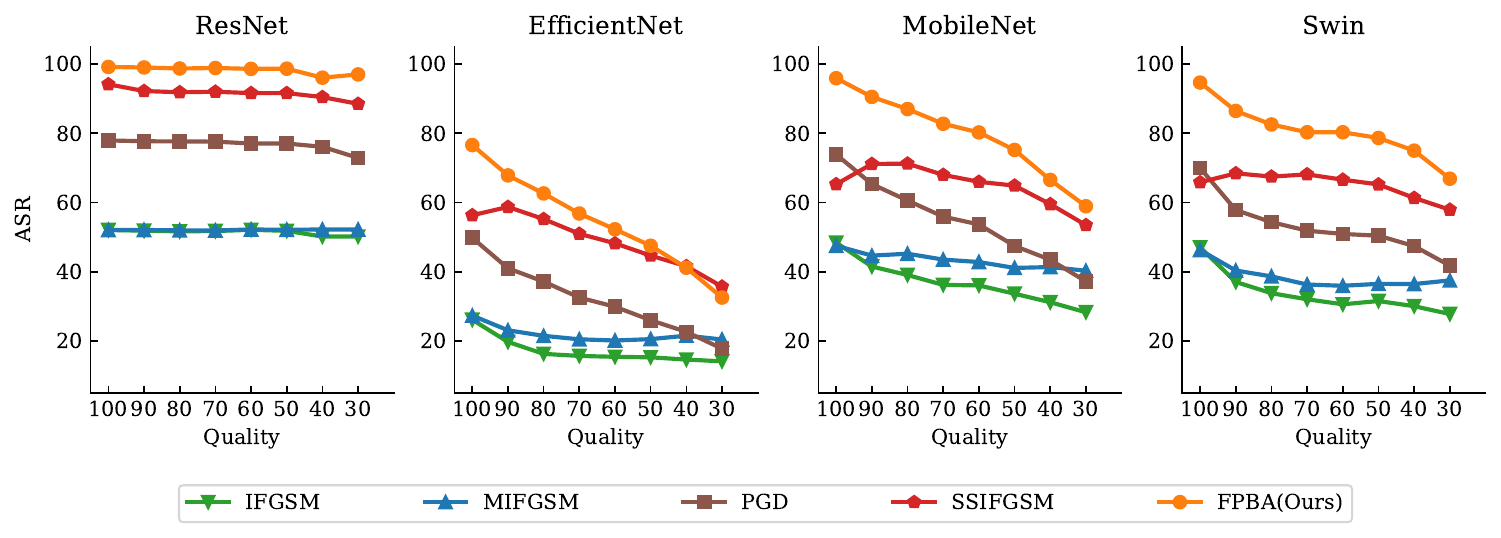}
  \caption{\revisionb{Attack performance on ResNet, Efficient, MobileNet and Swin against JPEG compression on  Synthetic LSUN(ProGAN) dataset. Adversarial samples are crafted from ResNet.}}
  \label{fig:jpeg}
  \vspace{-0.5cm}
\end{figure*} 

Since AT is not a reliable defense method for AIGI detection, we need to explore alternative defense strategies. Considering that AIGI detection typically~\cite{wang2020cnn} uses JPEG compression and Gaussian blurring as data preprocessing during training to improve their robustness, we follow the suggestions from~\cite{wang2020cnn, zhu2024genimage} to utilize Gaussian blurring and JPEG compression as a defense. In addition, images are often compressed during propagation, which may also distort adversarial noise patterns in real-world scenarios. Therefore, ensuring the effectiveness of adversarial attacks on compressed images is also crucial for simulating real-world scenarios. To this end, we perform JPEG compression with quality factors of 30, 40, 50, 60, 70, 80, 90 and 100 to the tested images. We evaluate the degradation in attack performance using four models: ResNet-50, EfficientNet, MobileNet and Swin Transformer. All adversarial samples are crafted from ResNet-50. The results are reported in~\cref{fig:jpeg}. For white-box attacks on ResNet, JPEG compression has a negligible effect on the attack success rate across different attack methods, maintaining top-1 attack performance. For transfer-based attacks, the attack strength on victim models decreases as the quality factor decreases. However, FPBA consistently maintains the highest attack success rate, demonstrating the effectiveness of our attack method under real-world scenarios.

Next, we follow the suggestions from~\cite{wang2020cnn, zhu2024genimage} to simultaneously utilize Gaussian blurring and JPEG compression as a defense. As shown in \cref{tab:ad}, existing attack methods all suffer from performance degradation as the probability of using Gaussian blurring and JPEG compression increases during training, while FPBA still contains a high attack success rate, which validates its effectiveness.

\begin{table}[!tb]
\caption{Defense models(Blur+JPEG) against adversarial attack on ProGAN Dataset. Blur+JPEG(0.1) and Blur+JPEG(0.5) represented that images are blurred and JPEG-ed with 10\% and 50\% probability respectively. We report the attack success rate(\%).}
\vspace{-0.5cm}
\label{tab:ad}
\begin{center}
    \setlength{\tabcolsep}{3pt}
    \renewcommand\arraystretch{0.99}	
    \scalebox{0.96}{%
\begin{tabular}{l|cc|cc}
\hline
\multicolumn{1}{c|}{} &  \multicolumn{2}{c}{\textbf{Swin-ViT}} \\ \cline{2-3}
    & \multicolumn{1}{c|}{Blur+JPEG(0.1)} & \multicolumn{1}{c}{Blur+JPEG(0.5)}\\
\hline
IFGSM    & 98.8 & \multicolumn{1}{c}{86.6(-12.2)} \\
MIFGSM   & 98.3 & \multicolumn{1}{c}{92.7(-5.6)} \\
PGD      & 99.3 & \multicolumn{1}{c}{87.3(-12)} \\
S$^2$I     & 100.0 & \multicolumn{1}{c}{80.9(-19.1)} \\
FPBA     & 100.0 & \multicolumn{1}{c}{99.2(-0.8)} \\
\hline
\end{tabular}}
\end{center}
\vspace{-0.3cm}
\end{table}

\begin{table*}[!htb]

  \caption{The attack success rate(\%) on CNN-based, ViT-based and Frequency-based models on the Synthetic FFHQ datasets. “Average” was calculated as the average transfer success rate over all victim models except for the surrogate model. We mark the white-box attack results in gray, and black-box attack results are not marked with colors.}
      \centering
  \label{tab:deepfake_dataset}
  \vspace{-0.3cm}
      \setlength{\tabcolsep}{3pt}
    \renewcommand\arraystretch{0.95}	
    \scalebox{1}{%
  \begin{tabular}{c|c|ccccccccc|c}
   \hline
 & \textbf{Surrogate Model} & \multicolumn{1}{c|}{\textbf{Attack Methods}} & \textbf{CNNSpot} & \textbf{DenseNet} & \textbf{EfficientNet} & \textbf{MobileNet} & \textbf{Spec} & \textbf{DCTA} & \textbf{ViT} & \textbf{Swin} & \textbf{Average}\\ 
     \hline

\multicolumn{1}{c|}{} & \multicolumn{1}{c|}{}  & \multicolumn{1}{c|}{IFGSM} & \colorbox{mygray}{67.5} & 67.5 & 55.3 & 67.5 & 28.9 & 32.5 & 16.2 & 17.5 & 40.8\\
\multicolumn{1}{c|}{} & \multicolumn{1}{c|}{}  & \multicolumn{1}{c|}{MIFGSM} & \colorbox{mygray}{67.5} & 67.5 & 54.9 & 67.6 & 28.8 & 32.6 & 20.7 & 17.9 & 41.4\\
\multicolumn{1}{c|}{} & \multicolumn{1}{c|}{} & \multicolumn{1}{c|}{PGD} & \colorbox{mygray}{\textbf{100}} & \textbf{100} & \textbf{88.6} & 99.9 & \textbf{49.9} & 50 & 40.7 & 50.0 & 68.4\\
\multicolumn{1}{c|}{} & \multicolumn{1}{c|}{CNNSpot}  & \multicolumn{1}{c|}{S$^2$I} & \colorbox{mygray}{\textbf{100}} & \textbf{100} & 77.7 & 90.1 & 46.7 & \textbf{52.9} & 37.6 & \textbf{55.8} & 65.8\\

\multicolumn{1}{c|}{} & \multicolumn{1}{c|}{} & \multicolumn{1}{c|}{\revisionb{SSAH}} & \colorbox{mygray}{\revisionb{99.0}} & \revisionb{54.5} & \revisionb{44.1} & \revisionb{41.0} & \revisionb{2.6} & \revisionb{7.2} & \revisionb{1.0} & \revisionb{4.2} & \revisionb{22.1}\\

\multicolumn{1}{c|}{} & \multicolumn{1}{c|}{} & \multicolumn{1}{c|}{\textbf{FPBA(Ours)}}   & \colorbox{mygray}{\textbf{100}} & \textbf{100} & 86.8 & \textbf{100} & 49.4 & 49.6 & \textbf{43.1} & 50.9 & \textbf{68.5}\\  \cline{2-12}

\multicolumn{1}{c|}{} & \multicolumn{1}{c|}{} & \multicolumn{1}{c|}{IFGSM} & 10.7 & 12.7 & 23.8 & \colorbox{mygray}{60.4} & 24.5 & 28.4 & 7.3 & 10.2 & 16.8\\
\multicolumn{1}{c|}{} & \multicolumn{1}{c|}{} & \multicolumn{1}{c|}{MIFGSM} & 10.3 & 12.6 & 23.2 & \colorbox{mygray}{60.4} & 25.4 & 28.7 & 9.1 & 10.2 & 17.1\\
\multicolumn{1}{c|}{} & \multicolumn{1}{c|}{} & \multicolumn{1}{c|}{PGD} & 49.8 & 50.2 & 57.5 & \colorbox{mygray}{\textbf{100}} & 49.7 & 49.8 & 38.8 & 49.8 & 49.4\\
\multicolumn{1}{c|}{} & \multicolumn{1}{c|}{MobileNet} & \multicolumn{1}{c|}{S$^2$I} & \textbf{82.8} & \textbf{86.7} & \textbf{78.5} & \colorbox{mygray}{\textbf{100}} & \textbf{51.4} & \textbf{51.8} & 28.9 & 49.7 & \textbf{61.4}\\

\multicolumn{1}{c|}{} & \multicolumn{1}{c|}{} & \multicolumn{1}{c|}{\revisionb{SSAH}} & \revisionb{1.9} & \revisionb{36.4} & \revisionb{43.7} & \colorbox{mygray}{\revisionb{98.5}} & \revisionb{2.0} & \revisionb{6.2} & \revisionb{0.8} & \revisionb{2.4} & \revisionb{13.3}\\

\multicolumn{1}{c|}{\multirow{-12}{*}{\rotatebox{90}{FFHQ (Style GAN)}}} & \multicolumn{1}{c|}{} & \multicolumn{1}{c|}{\textbf{FPBA(Ours)}}  & 51.2 & 55.2 & 66.6 & \colorbox{mygray}{\textbf{100}} & 50.3 & 49.9 & \textbf{39.4} & \textbf{49.8} & 51.8 \\  
 
\hline
\end{tabular}}
\vspace{-0.5cm}
\end{table*}

\subsection{Visual Quality Analysis}

\begin{table}[!htb]
\caption{The visual quality of different attack samples in terms of the average MSE, PSNR and SSIM scores.}
\vspace{-0.3cm}
\centering
\label{tab:image_quality}
\begin{tabular}{l|ccc}
\hline
\textbf{Attack}           & \textbf{MSE$\downarrow$} & \textbf{PSNR(db)$\uparrow$} & \textbf{SSIM$\uparrow$} \\ \hline
PGD        & 30.00    & 33.51            & 0.88     \\
FakePolisher        & 34.63    & 32.92            & 0.88     \\
FPBA(Ours) & \textbf{16.08}    & \textbf{36.26}   & \textbf{0.94}  \\ \hline
\end{tabular}
\vspace{-0.2cm}
\end{table}

\begin{figure}[!htb]
  \centering
  \includegraphics[width=1\linewidth]{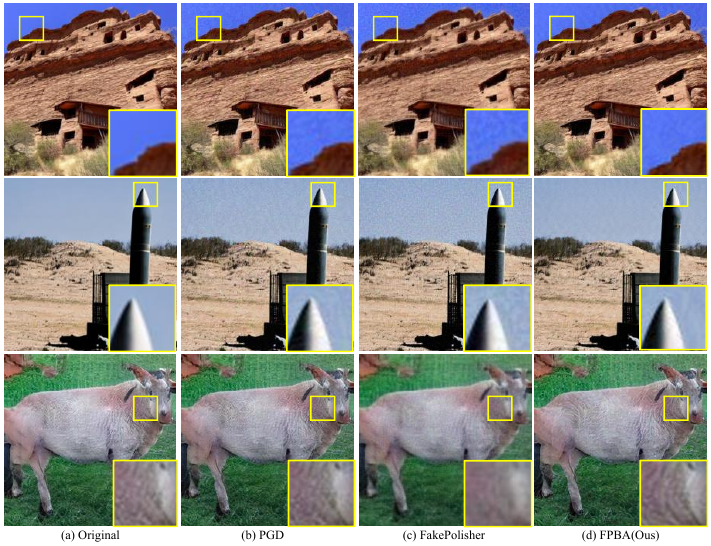}
  \vspace{-0.5cm}
  \caption{\revision{Visual comparison with PGD. (a): the original image generated by diffusion models. (b): adversarial examples crafted by PGD. (c): the original image generated by FakePolisher. (d): adversarial examples crafted by FPBA. The image quality of the adversarial example crafted by our method is much closer to the original image.}}
  \label{fig:vis}
\end{figure} 

To demonstrate the superior image quality achieved by our method, we conduct both qualitative and quantitative assessments. We first visualize adversarial examples generated by PGD, FakePolisher and our proposed FPBA in \cref{fig:vis}. Adversarial examples from PGD and FakePolisher show noticeable noise patterns upon zooming in. In addition, the fine-grained details of objects generated by FakePolisher are blurred, which is very obvious in the gota's texture and outline in its adversarial images. In contrast, FPBA generates more natural-looking adversarial examples. Further, we also report the quantitative results using common metrics for image quality assessment, including MSE, PSNR(db) and SSIM. As reported in \cref{tab:image_quality}, FPBA outperforms other baselines across all quality assessments by a large margin. This suggests that FPBA adding adversarial noise in the frequency domain, rather than directly in the spatial domain, is more imperceptible to observers.

\subsection{Additional Performance Analysis}
\subsubsection{The Phenomenon of Gradient Masking in AIGI Detectors}
\label{sec:GV}
\begin{figure}[!htb]
  \centering
  \includegraphics[width=0.8\linewidth]{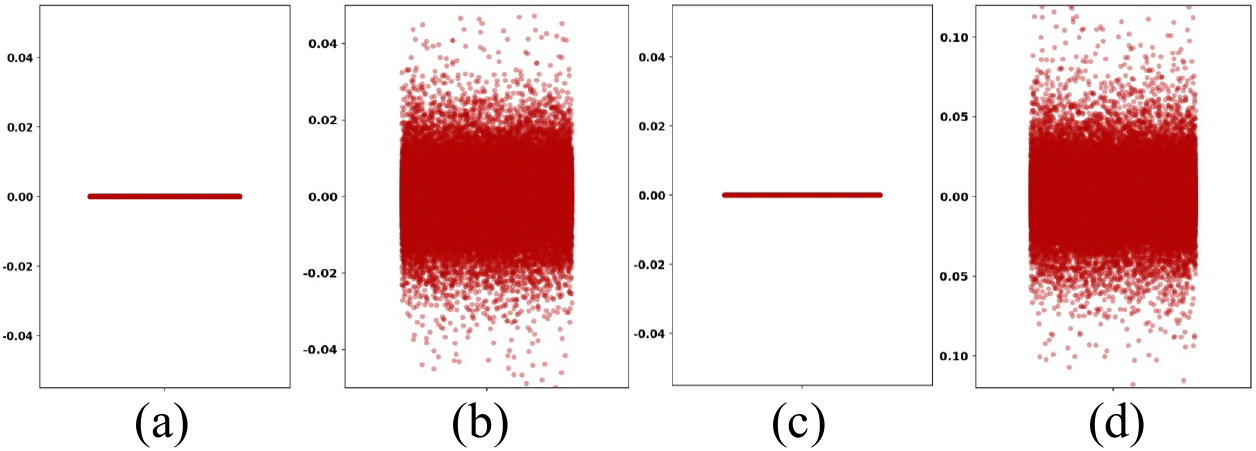}
  \vspace{-0.3cm}
  \caption{The gradient components of CNNSpot on adversarial examples generated by different attack methods. (a) gradient components of IFGSM; (b) gradient components of PGD; (c) gradient components of MIFGSM; (d) gradient components of MIFGSM with random initialization.}
  \label{fig:gradient}
\end{figure}

In \cref{tab:commands}, we find that the attack results of MIFGSM are similar to IFGSM, and significantly lower than PGD, in contrast to the common belief that adversarial examples generated by momentum iterative methods have a higher success rate~\cite{dong2018boosting}. It indicates the possibility of gradient masking~\cite{athalye2018obfuscated}. To investigate whether AIGI detectors exist in gradient masking, we analyze the aggregated gradient convergence properties. We take the adversarial examples crafted by CNNSpot on ProGan as an example. For each plot, we randomly sample 500 adversarial examples generated by a specific attack to compute their expected loss gradients. Each dot shown in \cref{fig:gradient} represents a component of the expected loss gradient from each image, in which there are a total of 75k loss gradient components. 

Most gradient components of adversarial examples generated by IFGSM (\cref{fig:gradient}(a)) and MIFGSM (\cref{fig:gradient} (c)) tend to stabilize around zero, indicating the vanishing gradient leading to a limited transferability to IFGSM and MIFGSM. Because the only difference between IFGSM and PGD is that PGD randomly chooses the starting point within the $l_{\infty}$ constraint, we apply random initialization to MIFGSM and find the value of gradient component increase (\cref{fig:gradient} (d)), w.r.t. the average success rate increasing from 40.4\% to 58.6\% (still lower 10.6\% than ours). This analysis formally demonstrates the phenomenon of gradient masking in AIGI detectors. Therefore, we advocate for future work for attacks on AIGI detectors to employ randomized-based strategies to circumvent the effect of gradient masking. Our proposed method conducts the spectrum transformation in the frequency domain and hence is also effective for gradient masking.   

\subsubsection{Evaluation on Deepfake Datasets}
As we mentioned before, there is a big difference between AIGI detection and Deepfake detection. But we empirically found that FPBA is also effective for face forgery detection. To demonstrate this, we conduct the experiments on synthetic FFHQ face datasets~\cite{shamshad2023evading}, which consists of 50k real face images from FFHQ and 50k fake generated face images created by StyleGAN2~\cite{ffhq}. We report the results in \cref{tab:deepfake_dataset}. Although FPBA is not specifically designed for Deepfake detection, it still achieves the best white-box attack performance. For transfer-based attack, FPBA also has the top-2 performance, demonstrating that FPBA is a universal threat across both AIGI and Deepfake detection. 
\begin{table*}[t]
\centering
\caption{\revision{Attack success rate (\%) under different frequency transforms on the ProGAN dataset.}}
\vspace{-0.3cm}
\resizebox{0.8\linewidth}{!}{
\begin{tabular}{ccccccccccc}
\hline
\textbf{\revision{Surrogate}} & \revision{\textbf{Transform}} & \revision{\textbf{CNNSpot}} & \revision{\textbf{DenseNet}} & \revision{\textbf{EfficientNet}} & \revision{\textbf{MobileNet}} & \revision{\textbf{Spec}} & \revision{\textbf{DCTA}} & \revision{\textbf{ViT}} & \revision{\textbf{Swin}} & \revision{\textbf{Avg}} \\
\hline
\multirow{3}{*}{\revision{CNNSpot}} 
& \revision{FFT}     & \revision{98.75} & \revision{98.15} & \revision{76.05} & \revision{95.70} & \revision{19.60} & \revision{47.40} & \revision{50.85} & \revision{94.45} & \revision{72.62} \\
&\revision{Wavelet} & \revision{99.20} & \revision{98.10} & \revision{71.20} & \revision{94.55} & \revision{19.30} & \revision{46.90} & \revision{50.60} & \revision{92.65} & \revision{71.56} \\
& \revision{DCT}  & \revision{98.90} & \revision{98.00} & \revision{76.40} & \revision{95.90} & \revision{19.80} & \revision{48.00} & \revision{51.50} & \revision{94.80} & \revision{\textbf{72.91}} \\
\hline
\multirow{3}{*}{\revision{MobileNet}} 
& \revision{FFT}     & \revision{31.65} & \revision{40.00} & \revision{51.00} & \revision{99.50} & \revision{22.35} & \revision{36.50} & \revision{12.45} & \revision{44.80} & \revision{42.28} \\
& \revision{Wavelet} & \revision{23.04} & \revision{31.15} & \revision{32.04} & \revision{91.40} & \revision{21.85} & \revision{36.45} & \revision{11.20} & \revision{36.55} & \revision{35.46} \\
& \revision{DCT}  & \revision{32.20} & \revision{40.40} & \revision{51.70} & \revision{99.60} & \revision{22.60} & \revision{36.40} & \revision{12.30} & \revision{44.60} & \revision{\textbf{42.48}}\\
\hline
\end{tabular}}
\label{tab:freq_ablation}
\vspace{-0.3cm}
\end{table*}

\begin{table}[!tb]
\caption{Ablation Study on ProGAN dataset. The adversarial samples are crafted from CNNSpot.}
\label{tab:ablation_diff_atk_patterns}
\vspace{-0.4cm}
\begin{center}
    \setlength{\tabcolsep}{3pt}
    \renewcommand\arraystretch{0.99}	
    \scalebox{1.1}{%
\begin{tabular}{l|cccccc}
\hline
\textbf{Attack}  & \textbf{DenseNet} & \textbf{MobileNet} &  \textbf{Spec}  & \textbf{ViT}  \\
\hline
Spatial &  71.5  &  34.0 & 27.1  & 11.0  \\
Frequency   &  91.0  &  64.7 & 36.7  & 16.9  \\
Spatial-frequency  &  96.2  &  66.2 & 47.9  & 21.0 \\
\hline
\end{tabular}}
\end{center}
\vspace{-0.5cm}
\end{table}

\begin{table}[!t]
\caption{The attack success rates (\%) of FPBA on normally trained detectors w.r.t the number $N$ of spectrum transformations. “Average” was calculated as the average transfer success rate over all victim models except for the surrogate model.} 
\label{tab:exper_N_number}
\vspace{-0.4cm}
\begin{center}
    \setlength{\tabcolsep}{3pt}
    \renewcommand\arraystretch{0.99}	
    \scalebox{0.95 }{%
\begin{tabular}{c|ccccccccc}
\hline
\textbf{$N$} & \textbf{ResNet} & \textbf{DenNet} & \textbf{EffNet} & \textbf{MobNet}  & \textbf{Spec} & \textbf{DCTA} & \textbf{ViT} & \textbf{Swin} & \textbf{Ave.}\\
\hline
0   & 64.5 & 64.0 & 37.3 & 60.6 & 17.5 & 48.4 & 40.5 & 59.7 & 48.8\\ 
5   & 98.9 & 98.0 & 76.4 & 95.9 & 19.8 & 48.0 & 51.5 & 94.8 & 69.2\\ 
10  & 99.6 & 99.1 & 78.6 & 97.5 & 19.9 & 48.0 & 52.4 & 96.5 & 70.3\\ 
15  & 99.4 & 99.0 & 78.2 & 96.6 & 20.2 & 48.3 & 53.0 & 96.2 & 70.2\\ 
20  & 99.2 & 98.7 & 77.5 & 96.7 & 20.7 & 48.6 & 51.1 & 95.6 & 69.8\\ 
\hline
\end{tabular}}
\end{center}
\vspace{-0.3cm}
\end{table}

\subsection{Ablation Study}
\paragraph{\revision{The Selection of Frequency Transforms}}
\revision{We conduct an ablation study comparing the performance of different frequency transforms, including Fast Fourier Transform (FFT), Wavelet Transform (Wavelet) and Discrete Cosine Transform (DCT). The results in \cref{tab:freq_ablation} show that FFT, Walevet and DCT all show high attack performance, with DWT performing the best. So we use DCT by default.}

\paragraph{Spatial and Frequency Attack}
We conduct an ablation study in \cref{tab:ablation_diff_atk_patterns} to investigate the impact of computing the gradient in different domains. In comparison with computing the attack gradient solely in the spatial domain or frequency domain, our spatial-frequency attack achieves higher transferability. More ablation studies can be found in the supplementary material.

\paragraph{Number ($N$) of Spectrum Transformations}
In this study, we investigate the impacts on the number ($N$) of spectrum transformations, which can simulate diverse substitute models. The adversarial examples were crafted from ResNet-50 in the Synthetic LSUN(ProGAN) dataset. With the exception of $N$, other hyperparameters keep the same with the default settings in the paper. As shown in \cref{tab:exper_N_number}, with $N$ increasing 5 from 0, FPBA shows a substantial augmentation in attack success rate. Further, when $N>5$, there is a diminishing gain in attack success rate but with increased computation. Considering the trade-off between attack performance and computation cost, we set $N=5$ by default.

\section{Conclusion}
In this paper, we investigate the adversarial robustness of AIGI detectors and propose a novel frequency-based post-train Bayesian attack that extends the Bayesian attack family. Through extensive experiments across models, generators, and defenses under both white-box and black-box settings, we draw the following conclusions: (1) State-of-the-art AIGI detectors remain highly vulnerable to adversarial attacks. Although adversarial training is effective in other domains, it cannot be directly applied in AIGI detection, posing major challenges for building robust detectors. (2) Many detectors rely on gradient masking to hinder the transferability of gradient-based attacks, yet the defense is easily bypassed with simple random initialization. (3) Attack success correlates positively with detection accuracy, suggesting that robustness evaluation on low-accuracy subsets is uninformative. Overall, our findings highlight that constructing adversarially robust AIGI detectors remains an open problem. We hope our work serves as both a benchmark and a call to further research in this critical area.

\section*{Acknowledgments}
This work was supported in part by the NSF China (No. 62302139, 62576020) and Fundamental Research Funds for the Central Universities of China (PA2025IISL0113, JZ2025HGTB0227).

\bibliographystyle{IEEEtran}
\bibliography{Manuscript, supplementary/supp}

\begin{IEEEbiography}[{\includegraphics[width=1in,height=1.25in,clip,keepaspectratio]{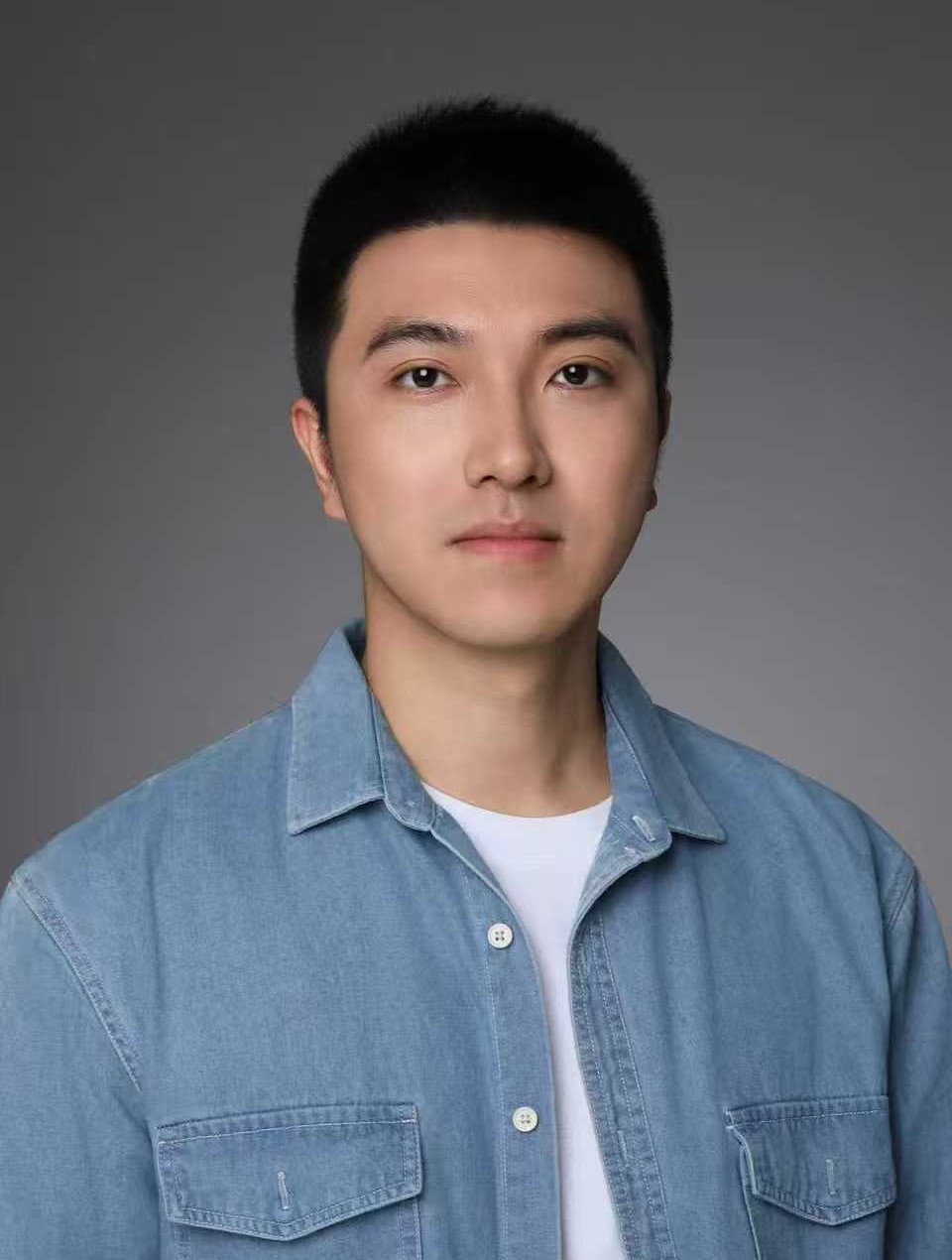}}]{Yunfeng Diao} is a Lecturer in the School of Computer Science and Information Engineering, Hefei University of Technology, China. He received his PhD from Southwest Jiaotong University, China. His current research interests include computer vision and the security of machine learning. He has published over 30 papers in leading venues such as CVPR, ICLR, ICCV, ICML, IEEE TMM and IEEE TCSVT. He has been an editorial board member of IJAACS, the chief organizer of several workshops at IJCAI, also regularly reviewing for top-tier journals and conferences. He has received the Outstanding Contribution Award at the IJCAI Workshop and the Best Paper Award at the IROS RODGE Workshop.
\end{IEEEbiography}
\begin{IEEEbiography}[{\includegraphics[width=1in,height=1.25in,clip,keepaspectratio]{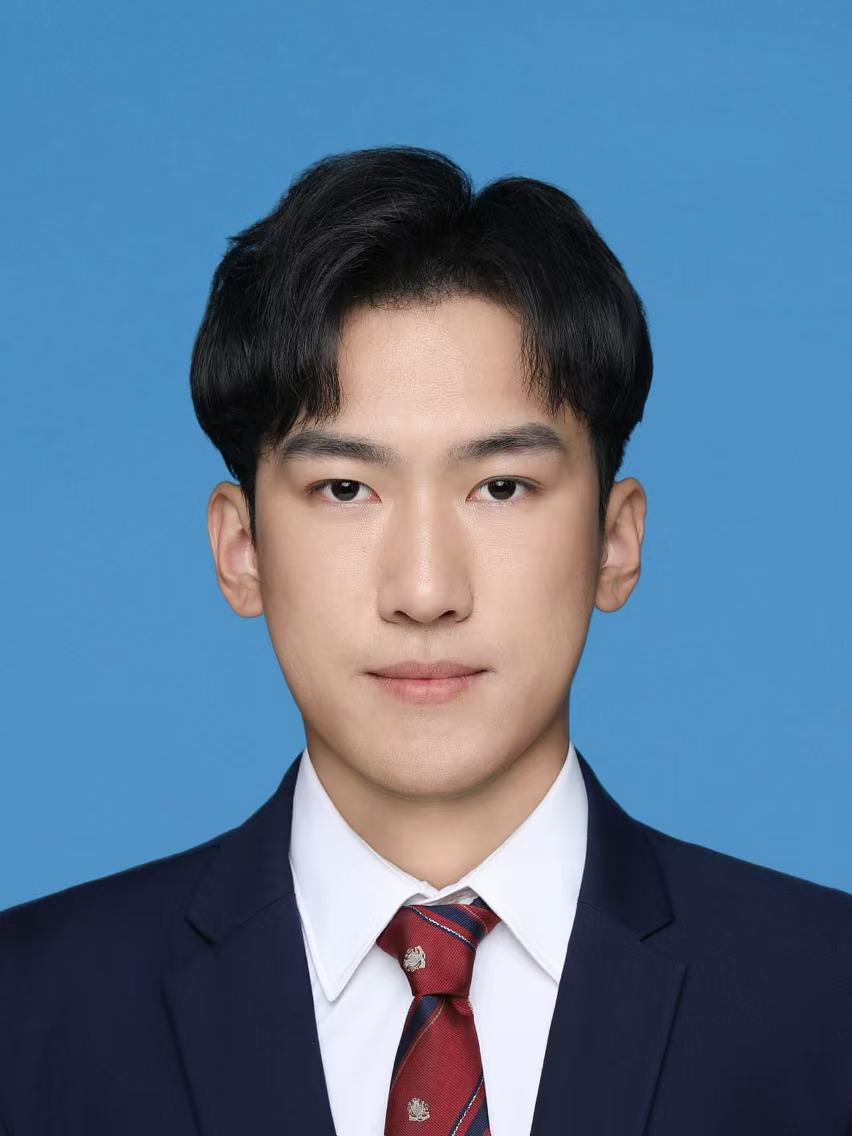}}]{Naixin Zhai} received the B.S. degree from Hefei University of Technology, Hefei, China, in 2024. He is current working toward the M.S. degree  with the Department of Electronic Engineering and Information Science. His research interests include deep learning and synthetic image detection.
\end{IEEEbiography}
\begin{IEEEbiography}[{\includegraphics[width=1in,height=1.25in,clip,keepaspectratio]{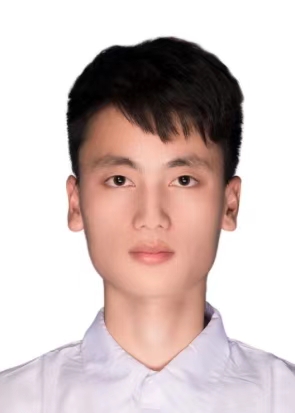}}]{Changtao Miao} received his B.S. degree in 2019 from AnHui University. He is currently pursuing the Ph.D. degree in Cyber Science and Technology in University of Science and Technology of China. His research interests include face forgery forensics and face manipulation.
\end{IEEEbiography}
\begin{IEEEbiography}[{\includegraphics[width=1in,height=1.25in,clip,keepaspectratio]{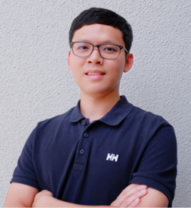}}]{Zitong Yu} (Senior Member, IEEE) received the Ph.D. degree in computer science and engineering from the University of Oulu, Finland, in 2022. Currently, he is an Assistant Professor with Great Bay University, China. He was a Post-Doctoral Researcher with the ROSE Laboratory, Nanyang Technological University. He was a Visiting Scholar with TVG, University of Oxford, from July 2021 to November 2021. His research interests include human-centric computer vision and biometric security. He was a recipient of IAPR Best Student Paper Award, IEEE Finland Section Best Student Conference Paper Award 2020.
\end{IEEEbiography}
\begin{IEEEbiography}[{\includegraphics[width=1in,height=1.25in,clip,keepaspectratio]{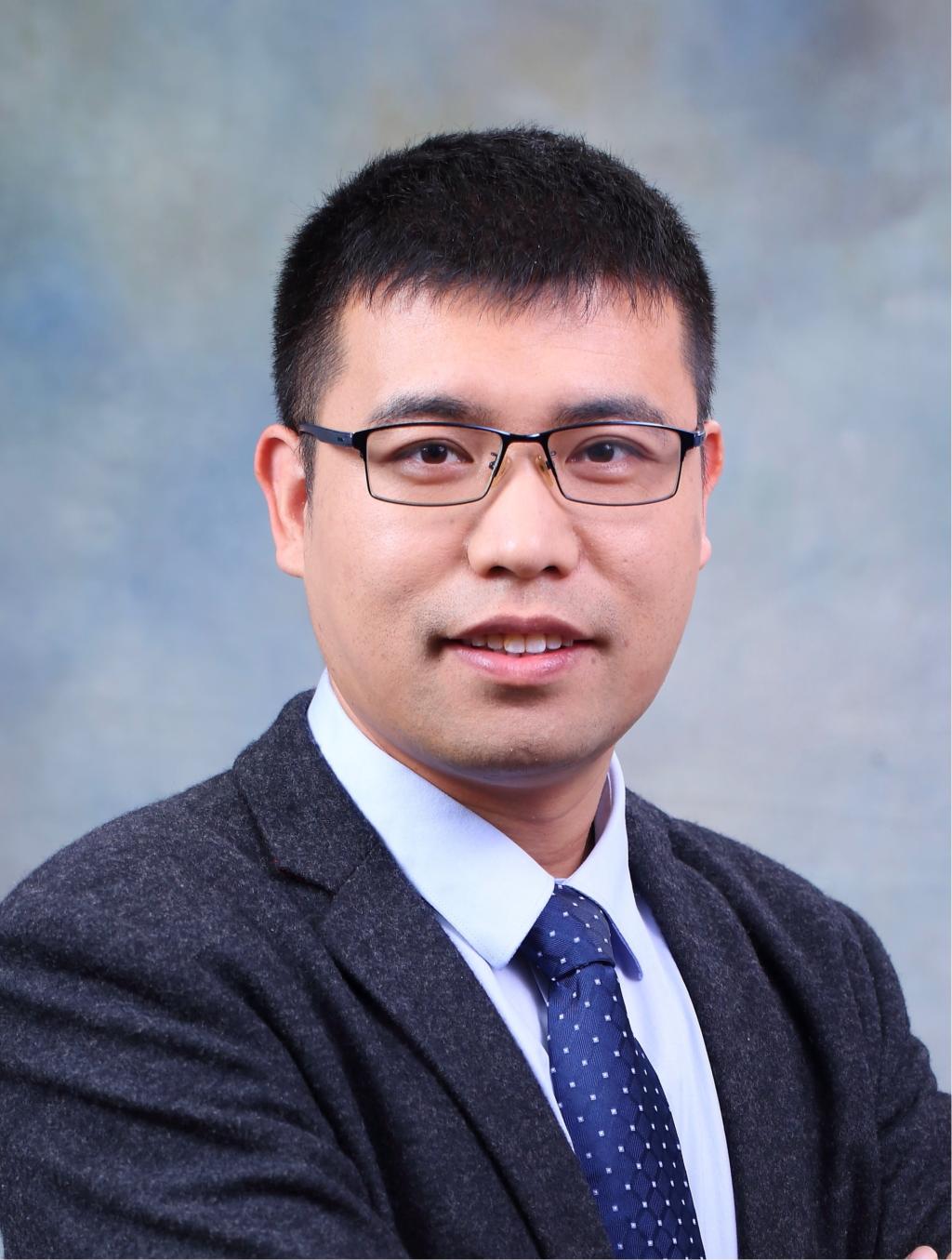}}]{Xingxing Wei} received his Ph.D degree in computer science from Tianjin University, and B.S. degree in Automation from Beihang University (BUAA), China. He is now an Associate Professor at Beihang University (BUAA). His research interests include computer vision, adversarial machine learning and its applications to multimedia content analysis. He is the author of referred journals and conferences in IEEE TPAMI, TMM, TCYB, TGRS, IJCV, PR, CVIU, CVPR, ICCV, ECCV, ACMMM, AAAI, IJCAI etc.
\end{IEEEbiography}
\vspace{-33pt}
\begin{IEEEbiography}[{\includegraphics[width=1in,height=1.25in,clip,keepaspectratio]{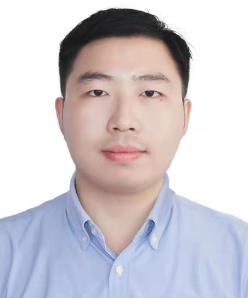}}]{Xun Yang} (Senior Member, IEEE) received the Ph.D. degree from the Hefei University of Technology, Hefei, China, in 2017. He is currently a Professor with the Department of Electronic Engineering and Information Science, University of Science and Technology of China (USTC). From 2015 to 2017, he visited the University of Technology Sydney (UTS), Australia, as a Joint Ph.D. Student. He was a Research Fellow with the NExT++ Research Center, National University of Singapore (NUS), from 2018 to 2021. His current research interests include information retrieval, cross-media analysis and reasoning, and computer vision. He serves as an Associate Editor for IEEE TRANSACTIONS ON BIG DATA and Multimedia Systems journal.
\end{IEEEbiography}
\begin{IEEEbiography}[{\includegraphics[width=1in,height=1.25in,clip,keepaspectratio]{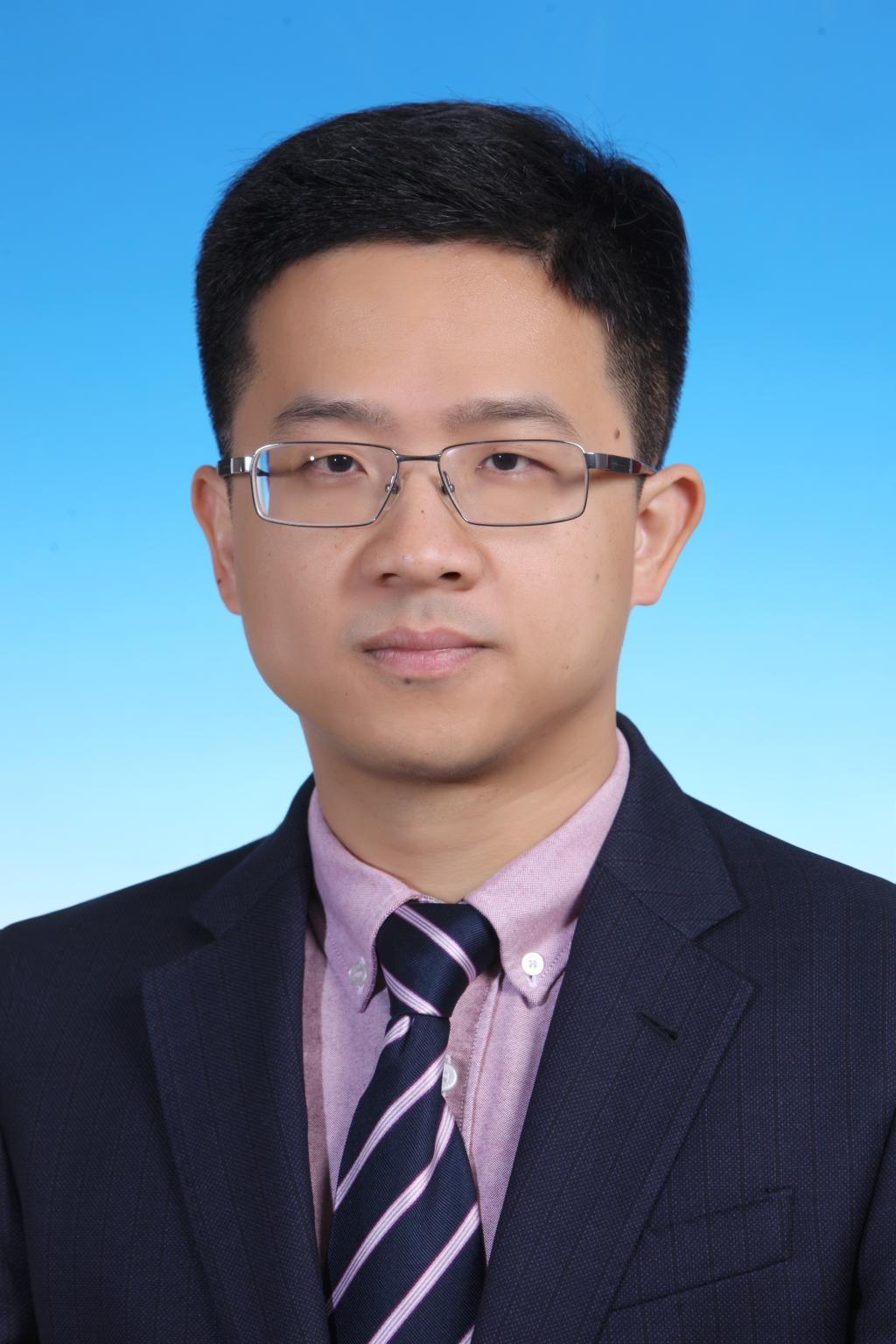}}]{Meng Wang} (Fellow, IEEE) is a professor at Hefei University of Technology, China. His current research interests include multimedia content analysis, computer vision, and pattern recognition. He received paper prizes or awards from ACM MM 2009 (Best Paper Award), ACM MM 2010 (Best Paper Award), MMM 2010 (Best Paper Award), SIGIR 2015 (Best Paper Honorable Mention), IEEE TMM 2015 and 2016 (Prize Paper Award Honorable Mention) and ACM TOMM 2018 (Nicolas D. Georganas Best Paper Award), etc. He currently serves on the editorial/advisory boards of IEEE TPAMI, IEEE TMM, IEEE TNNLS, etc. He is a fellow of IEEE and IAPR.
\end{IEEEbiography}

\clearpage
\title{Vulnerabilities in AI-generated Image Detection: \\The Challenge of Adversarial Attacks\\
---Supplementary Document---}

\maketitle

\begin{figure*}[b]
  \centering
    \scalebox{0.8}{
  \includegraphics[width=1\linewidth]{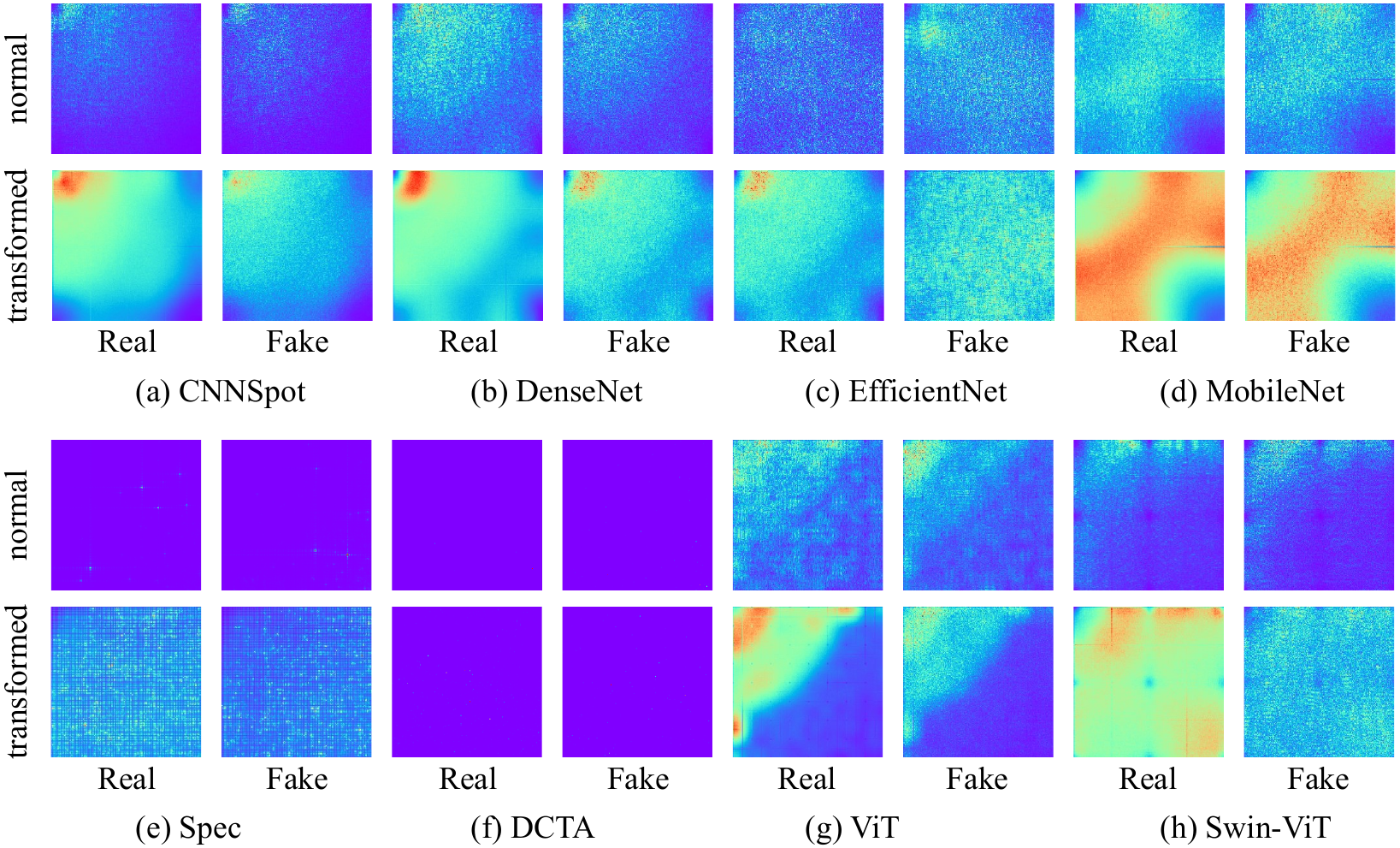}
  }
  \caption{Visualization of the spectrum saliency map with different models.}
  \label{fig:ssa_map}
\end{figure*}

\section{Additional Experiments}

\subsection{Experiment Details} 

We compared a series of distinguished backbone architectures pretrained on the ImageNet dataset. Specifically, we evaluated ResNet-50, DenseNet-121, EfficientNet-B4, MobileNet-V2, ViT-B/16, and Swin-B. Additionally, we examined Spec equipped with ResNet-34 architecture, as well as DCTA equipped with ResNet-50 architecture. In \cref{tab:exper_acc_detectors}, we enumerate the accuracy of all models on the test set as observed in our experiments. The meticulous evaluation of these foundational models not only revealed the potential vulnerabilities of AIGI detection models but also highlighted the adversarial threats they face. Our analytical results offer significant insights for understanding and improving the robustness of AIGI detection models.
\begin{table}[!ht]
\caption{The accuracy of each detector on different datasets.} 
\label{tab:exper_acc_detectors}
\vspace{-0.3cm}
\begin{center}
\resizebox{1\linewidth}{!}{
    \setlength{\tabcolsep}{3pt}
    \renewcommand\arraystretch{1.1}	
    \scalebox{1}{%
\begin{tabular}{c|cccccccc}
\hline
\textbf{Model} & \textbf{ResNet} & \textbf{DenNet} & \textbf{EffNet} & \textbf{MobNet}  & \textbf{Spec} & \textbf{DCTA} & \textbf{ViT} & \textbf{Swin} \\
\hline
LSUN(ProGAN)  & 99.99 & 99.91 & 99.96 & 99.93 & 99.90 & 99.91 & 99.74 & 100.00\\ 
GenImage(SD)  & 99.55 & 99.92 & 99.9 & 99.93 & 99.12 & 99.17 & 99.48 & 99.97 \\ 
FFHQ(Style GAN)  & 99.81 & 99.89 & 99.83 & 99.63 & 96.51 & 96.6 & 99.55 & 99.99 \\ 
\hline
\end{tabular}}}
\end{center}
\vspace{-0.3cm}
\end{table}

\subsection{\revision{Evaluation on ResNeXt}}
\revision{We use ResNext~\cite{xie2017aggregated} as the surrogate to evaluate the attack performance. As reported in \cref{tab:resnetx_data}, FPBA still achieves the best average attack success rate, which is consistent with other surrogate models.}
\begin{table}[!htb]
\label{tab:resnext}
\caption{\revision{Attack evaluation using ResNeXt as surrogate model on the ProGAN-LSUN dataset.}}
\vspace{-0.3cm}
\small
\label{tab:resnetx_data}
\begin{center}
\resizebox{1\linewidth}{!}{
    \setlength{\tabcolsep}{3pt}
    \renewcommand\arraystretch{1.1}	
    \scalebox{1}{%
\begin{tabular}{c|ccccccc|c}
\hline
\textbf{Attack} & \textbf{ResNeXt} & \textbf{ResNet} & \textbf{DenNet}   & \textbf{Spec} &\textbf{DCTA}  & \textbf{ViT}  & \textbf{Swin}   &\textbf{Ave.}\\
\hline
\textbf{MI-FGSM} &  54.2 & 29.7 & 21.3  & \underline{24.8} & \underline{49.7} & 14.2 & 6.6  & 28.6 \\
\textbf{PGD} & 77.1 & 35.4 & 30.9 & \textbf{28.0} & \textbf{50.4} & 20.6 & 28.0 & 38.6\\
\textbf{S$^2$I} &  98.6 & \textbf{90.6} & 86.9 & 13.4 & 36.9 & 15.0 & \textbf{68.0}& 58.5 \\
\textbf{FPBA} &  \textbf{99.9} & \underline{89.0} & \textbf{89.2} & 21.5 & 49.0 & \textbf{29.6} & \underline{64.1} & \textbf{63.2} \\
\hline
\end{tabular}}}
\end{center}
\vspace{0.3cm}
\end{table}

\subsection{Study of Attack Strength} 

\begin{figure*}[t]
  \centering
    \scalebox{0.8}{
  \includegraphics[width=1\linewidth]{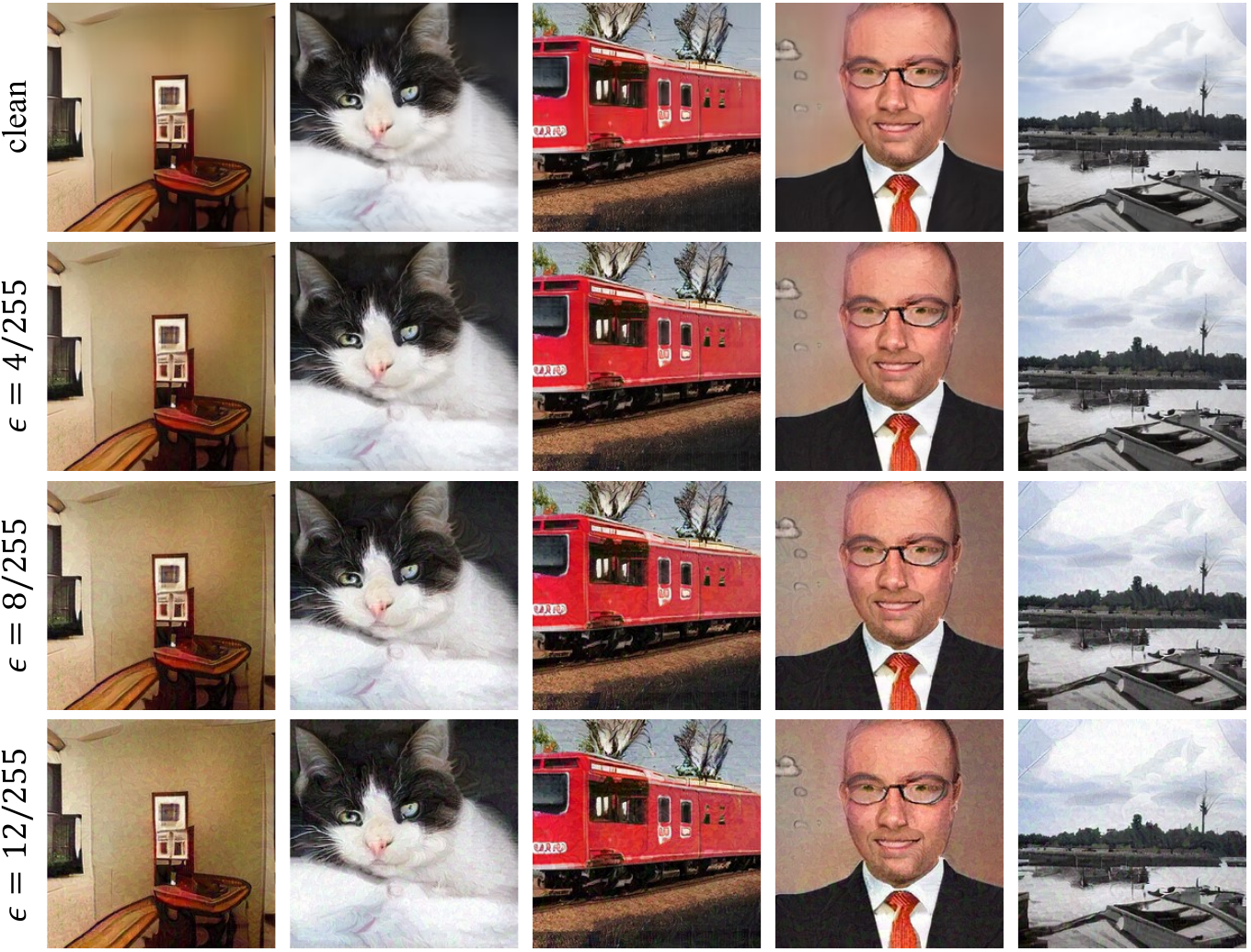}
  }
  \caption{The visual results of adversarial examples with different perturbation budgets.}
  \label{fig:eps4_8_12}
\end{figure*}

As shown in \cref{tab:exper_strength}, we report the comparison of the attack success rates of our FPBA with existing methods under different attack strengths. The results reflect a pattern consistent with when $\epsilon$=8/255, where our method can achieve globally optimal results compared to previous methods, and a larger perturbation budget consistently enhances adversarial transferability across various surrogate models. However, as the attack strength increases, the noise added causes more and more damage to the image, and the visual impact becomes greater, which is revealed in \cref{fig:eps4_8_12}.

\subsection{Spectrum Saliency Map} 

In \cref{fig:ssa_map}, we present the spectrum saliency map of all detectors for the original samples and the spectrum saliency map after frequency domain transformation.

\begin{table*}[!htb]
  \caption{The attack success rate(\%) on CNN-based, Vit-based and Frequency-based models on the Synthetic LSUN(ProGAN) with $\epsilon$=4/255, 8/255 or 12/255. “Average” was calculated as the average transfer success rate over all victim models except for the surrogate model.}
  \label{tab:exper_strength}
  \begin{center}
      \setlength{\tabcolsep}{3pt}
    \renewcommand\arraystretch{1.0}	
    \scalebox{1}{%
  \begin{tabular}{c|c|ccccccccc|c}
   \hline
 & \textbf{Surrogate Model} & \multicolumn{1}{c|}{\textbf{Attack Methods}} & \textbf{ResNet} & \textbf{DenNet} & \textbf{EffNet} & \textbf{MobNet} & \textbf{Spec} & \textbf{DCTA} & \textbf{ViT} & \textbf{Swin} & \textbf{Average}\\ 
     \hline

\multicolumn{1}{c|}{} & \multicolumn{1}{c|}{} & \multicolumn{1}{c|}{IFGSM}  & 52.1 & 44.6 & 14.0 & 36.8 & 15.0 & 41.2 & 12.9 & 24.6 & 27.0 \\
\multicolumn{1}{c|}{} & \multicolumn{1}{c|}{} & \multicolumn{1}{c|}{MIFGSM} & 52.1 & 50.0 & 23.5 & 45.4 & 15.0 & \textbf{45.1} & \textbf{22.9} & 38.3 & 34.3 \\
\multicolumn{1}{c|}{} & \multicolumn{1}{c|}{ResNet-50} & \multicolumn{1}{c|}{PGD} & 53.5 & 46.3 & 15.5 & 38.7 & \textbf{20.3} & 44.4 & 12.4 & 26.0 & 29.1 \\
\multicolumn{1}{c|}{} & \multicolumn{1}{c|}{} & \multicolumn{1}{c|}{S$^2$I} & 97.1 & 68.1 & 21.1 & 33.0 & 7.0 & 18.3 & 2.0 & 17.9 & 23.0 \\
\multicolumn{1}{c|}{\multirow{-5}{*}{\rotatebox{90}{$\epsilon$=4/255}}} & \multicolumn{1}{c|}{} & \multicolumn{1}{c|}{\textbf{FPBA(Ours)}} & \textbf{98.9} & \textbf{93.7} & \textbf{55.0} & \textbf{79.8} & 15.8 & 40.8 & 18.6 & \textbf{75.1} & \textbf{54.1}\\  \cline{1-12}

\multicolumn{1}{c|}{} & \multicolumn{1}{c|}{} & \multicolumn{1}{c|}{IFGSM} & 52.1 & 51.3 & 24.6 & 48.4 & 17.7 & 48.0 & 35.2 & 46.2 & 38.7 \\
\multicolumn{1}{c|}{} & \multicolumn{1}{c|}{} & \multicolumn{1}{c|}{MIFGSM} & 52.1 & 51.5 & 27.9 & 47.6 & \textbf{26.1} & \textbf{49.8} & 37.5 & 42.7 & 40.4 \\
\multicolumn{1}{c|}{} & \multicolumn{1}{c|}{ResNet-50} & \multicolumn{1}{c|}{PGD} & 78.3 & 77.6 & 49.4 & 73 & 25.6 & 47.6 & 41.0 & 70.2 & 54.9 \\
\multicolumn{1}{c|}{} & \multicolumn{1}{c|}{} & \multicolumn{1}{c|}{S$^2$I} & 97.8 & 86.4 & 61.5 & 78.6 & 20.5 & 40.8 & 11.7 & 74.5 & 53.4 \\
\multicolumn{1}{c|}{\multirow{-5}{*}{\rotatebox{90}{$\epsilon$=8/255}}} & \multicolumn{1}{c|}{} & \multicolumn{1}{c|}{\textbf{FPBA(Ours)}} & \textbf{98.9} & \textbf{98.0} & \textbf{76.4} & \textbf{95.9} & 19.8 & 48 & \textbf{51.5} & \textbf{94.8} & \textbf{69.2}\\  \cline{1-12}

\multicolumn{1}{c|}{} & \multicolumn{1}{c|}{} & \multicolumn{1}{c|}{IFGSM} & 52.1 & 51.9 & 28.7 & 50.0 & 18.3 & 49.1 & 40.6 & 50.2 & 41.2 \\
\multicolumn{1}{c|}{} & \multicolumn{1}{c|}{} & \multicolumn{1}{c|}{MIFGSM} & 52.1 & 51.8 & 28.7 & 47.4 & \textbf{37.2} & \textbf{50.0} & 34.7 & 38.9 & 41.2 \\
\multicolumn{1}{c|}{} & \multicolumn{1}{c|}{ResNet-50} & \multicolumn{1}{c|}{PGD} & 96.3 & 96.1 & 67.0 & 91.9 & 30.6 & 48.8 & 58.6 & 87.5 & 68.6 \\
\multicolumn{1}{c|}{} & \multicolumn{1}{c|}{} & \multicolumn{1}{c|}{S$^2$I} & 97.3 & 88.4 & 78.5 & 89.3 & 28.5 & 47.6 & 25.4 & 85.3 & 63.3 \\
\multicolumn{1}{c|}{\multirow{-5}{*}{\rotatebox{90}{$\epsilon$=12/255}}} & \multicolumn{1}{c|}{} & \multicolumn{1}{c|}{\textbf{FPBA(Ours)}}   & \textbf{98.9} & \textbf{99.3} & \textbf{82.9} & \textbf{98.5} & 20.7 & 49.2 & \textbf{68.7} & \textbf{98.0} & \textbf{73.9} \\  \cline{1-12}

\multicolumn{1}{c|}{} & \multicolumn{1}{c|}{} & \multicolumn{1}{c|}{IFGSM} & 4.0 & 5.6 & 12.5 & 75.4 & 12.3 & 24.6 & 4.8 & 3.4 & 9.6 \\
\multicolumn{1}{c|}{} & \multicolumn{1}{c|}{} & \multicolumn{1}{c|}{MIFGSM} & \textbf{9.0} & 11.5 & 16.5 & 75.4 & 13.7 & \textbf{26.4} & \textbf{6.4} & \textbf{10.9} & 13.5 \\
\multicolumn{1}{c|}{} & \multicolumn{1}{c|}{MobileNet-V2} & \multicolumn{1}{c|}{PGD} & 3.4 & 5.0 & 11.8 & 87.7 & \textbf{15.1} & 25.7 & 4.5 & 3.2 & 9.8 \\
\multicolumn{1}{c|}{} & \multicolumn{1}{c|}{} & \multicolumn{1}{c|}{S$^2$I} & 6.0 & 7.0 & 19.2 & 93.6 & 5.9 & 8.0 & 1.2 & 5.6 & 7.6\\
\multicolumn{1}{c|}{\multirow{-5}{*}{\rotatebox{90}{$\epsilon$=4/255}}} & \multicolumn{1}{c|}{} & \multicolumn{1}{c|}{\textbf{FPBA(Ours)}} & 7.9 & \textbf{11.8} & \textbf{26.0} & \textbf{99.5} & 14.4 & 25.8 & 4.5 & 9.5 & \textbf{14.3} \\  \cline{1-12}

\multicolumn{1}{c|}{} & \multicolumn{1}{c|}{} & \multicolumn{1}{c|}{IFGSM} & 14.3 & 17.8 & 22.4 & 75.4 & 17.2 & 34.9 & 9.2 & 18.4 & 19.2 \\
\multicolumn{1}{c|}{} & \multicolumn{1}{c|}{} & \multicolumn{1}{c|}{MIFGSM} & 23.8 & 23.9 & 28.6 & 75.4 & 19.6 & \textbf{42} & \textbf{15.5} & 24.8 & 25.5 \\
\multicolumn{1}{c|}{} & \multicolumn{1}{c|}{MobileNet-V2} & \multicolumn{1}{c|}{PGD} & 20.3 & 26.0 & 20.6 & 97.6 & \textbf{25.0} & 40.4 & 9.7 & 29.9 & 24.6 \\
\multicolumn{1}{c|}{} & \multicolumn{1}{c|}{} & \multicolumn{1}{c|}{S$^2$I} & 14.4 & 17.5 & 34.7 & 97.8 & 16.3 & 25.4 & 4.3 & 14.6 & 18.2 \\
\multicolumn{1}{c|}{\multirow{-5}{*}{\rotatebox{90}{$\epsilon$=8/255}}} & \multicolumn{1}{c|}{} & \multicolumn{1}{c|}{\textbf{FPBA(Ours)}}   & \textbf{32.2} & \textbf{40.4} & \textbf{51.7} & \textbf{99.6} & 22.6 & 36.4 & 12.3 & \textbf{44.6} & \textbf{34.3} \\  \cline{1-12}

\multicolumn{1}{c|}{} & \multicolumn{1}{c|}{} & \multicolumn{1}{c|}{IFGSM} & 18.8 & 21.4 & 27.7 & 75.4 & 18.0 & 40.2 & 11.4 & 23.0 & 22.9\\
\multicolumn{1}{c|}{} & \multicolumn{1}{c|}{} & \multicolumn{1}{c|}{MIFGSM} & 25.2 & 25.2 & 33.7 & 75.4 & 22.5 & \textbf{48.7} & \textbf{21.6} & 25.2 & 28.9\\
\multicolumn{1}{c|}{} & \multicolumn{1}{c|}{MobileNet-V2} & \multicolumn{1}{c|}{PGD} & 34.8 & 39.8 & 27.6 & 99.8 & \textbf{35.4} & 48.5 & 14.7 & 45.1 & 35.1\\
\multicolumn{1}{c|}{} & \multicolumn{1}{c|}{} & \multicolumn{1}{c|}{S$^2$I} & 28.1 & 33.5 & 44.0 & 98.6 & 23.3 & 34.5 & 7.9 & 35.5 & 29.5 \\
\multicolumn{1}{c|}{\multirow{-5}{*}{\rotatebox{90}{$\epsilon$=12/255}}} & \multicolumn{1}{c|}{} & \multicolumn{1}{c|}{\textbf{FPBA(Ours)}} & \textbf{42.3} & \textbf{45.7} & \textbf{56.3} & \textbf{99.9} & 27.0 & 42.0 & 18.7 & \textbf{49.0} & \textbf{40.1} \\  \cline{1-12}

\hline
\end{tabular}}
\end{center}
\end{table*}

\end{document}